\documentclass[lettersize,journal]{IEEEtran}
\usepackage{amsmath,amsfonts}
\usepackage{algorithmic}
\usepackage{algorithm}
\usepackage{array}
\usepackage{textcomp}
\usepackage{stfloats}
\usepackage{url}
\usepackage{verbatim}
\usepackage{graphicx}
\usepackage{subfigure}
\usepackage{booktabs}
\usepackage{cite}
\usepackage{pifont}
\usepackage{tikz}
\usepackage{pifont}
\usepackage{tabu}
\usepackage{multirow}
\usepackage{comment}
\usepackage{amsmath,amssymb}
\usepackage[perpage,symbol*]{footmisc}
\definecolor{applegreen}{rgb}{0.55, 0.71, 0.0}
\begin{document}

\title{Rethinking Cross-Domain Pedestrian Detection: A Background-Focused Distribution Alignment Framework for Instance-free One-Stage Detectors}

\author{Yancheng Cai, 
        Bo Zhang, 
        Baopu Li, \IEEEmembership{Member, IEEE},
        Tao Chen, \IEEEmembership{Senior Member, IEEE},
        Hongliang Yan,
        
        Jingdong Zhang,
        Jiahao Xu
        \vspace{-10pt}

\thanks{Yancheng Cai, Jingdong Zhang, Jiahao Xu, Tao Chen are with the School of
Information Science and Technology, Fudan University, Shanghai 200433,
China (Corresponding author: Tao Chen, e-mail: eetchen@fudan.edu.cn, tel: +86-2131242503).}
\thanks{Yancheng Cai was with Fudan when he finished this work, now he is with Department of Computer Science and Technology, University of Cambridge, CB3 0FD, UK (e-mail: yc613@cam.ac.uk).}
\thanks{Bo Zhang and Hongliang Yan are with Shanghai AI Laboratory, Shanghai, 200232, China (email: bo.zhangzx@gmail.com, yhldhit@gmail.com)}
\thanks{Baopu Li is Independent Researcher.}}

\newcommand{\zjd}[1]{\textcolor{orange}{#1}}



\maketitle

\begin{abstract}

Cross-domain pedestrian detection aims to generalize pedestrian detectors from one label-rich domain to another label-scarce domain, which is crucial for various real-world applications. Most recent works focus on domain alignment to train domain-adaptive detectors either at the instance level or image level. From a practical point of view, one-stage detectors are faster. Therefore, we concentrate on designing a cross-domain algorithm for rapid \footnote{Instance-free detectors refer to detectors that lack instance-level feature, such as the YOLO series and SSD. For detectors with available instance-level feature, like DETR~\cite{carion2020end}, they can directly perform instance-level domain adaptation, thus avoiding the foreground-background misalignment issue.} one-stage detectors that lacks instance-level proposals and can only perform image-level feature alignment. However, pure image-level feature alignment causes the foreground-background misalignment issue to arise, \textit{i.e.}, the foreground features in the source domain image are falsely aligned with background features in the target domain image. To address this issue, we systematically analyze the importance of foreground and background in image-level cross-domain alignment, and learn that background plays a more critical role in image-level cross-domain alignment. Therefore, we focus on cross-domain background feature alignment while minimizing the influence of foreground features on the cross-domain alignment stage. This paper proposes a novel framework, namely, background-focused distribution alignment (BFDA), to train domain adaptive  one-stage pedestrian detectors. Specifically, BFDA first decouples the background features from the whole image feature maps and then aligns them via a novel long-short-range discriminator. Extensive experiments demonstrate that compared to mainstream domain adaptation technologies, BFDA significantly enhances cross-domain pedestrian detection performance for either one-stage or two-stage detectors.  Moreover, by employing the efficient  one-stage detector (YOLOv5), BFDA can reach 217.4 FPS (640$\times$480 pixels) on NVIDIA Tesla V100 (7$\sim$12 times the FPS of the existing frameworks), which is highly significant for practical applications. The code from this study can be checked here: \textit{\textcolor{orange}{https://github.com/caiyancheng/BFDA}}.
\end{abstract}


\begin{figure*}
\centering
\vspace{-16pt}
\includegraphics[width=1\linewidth]{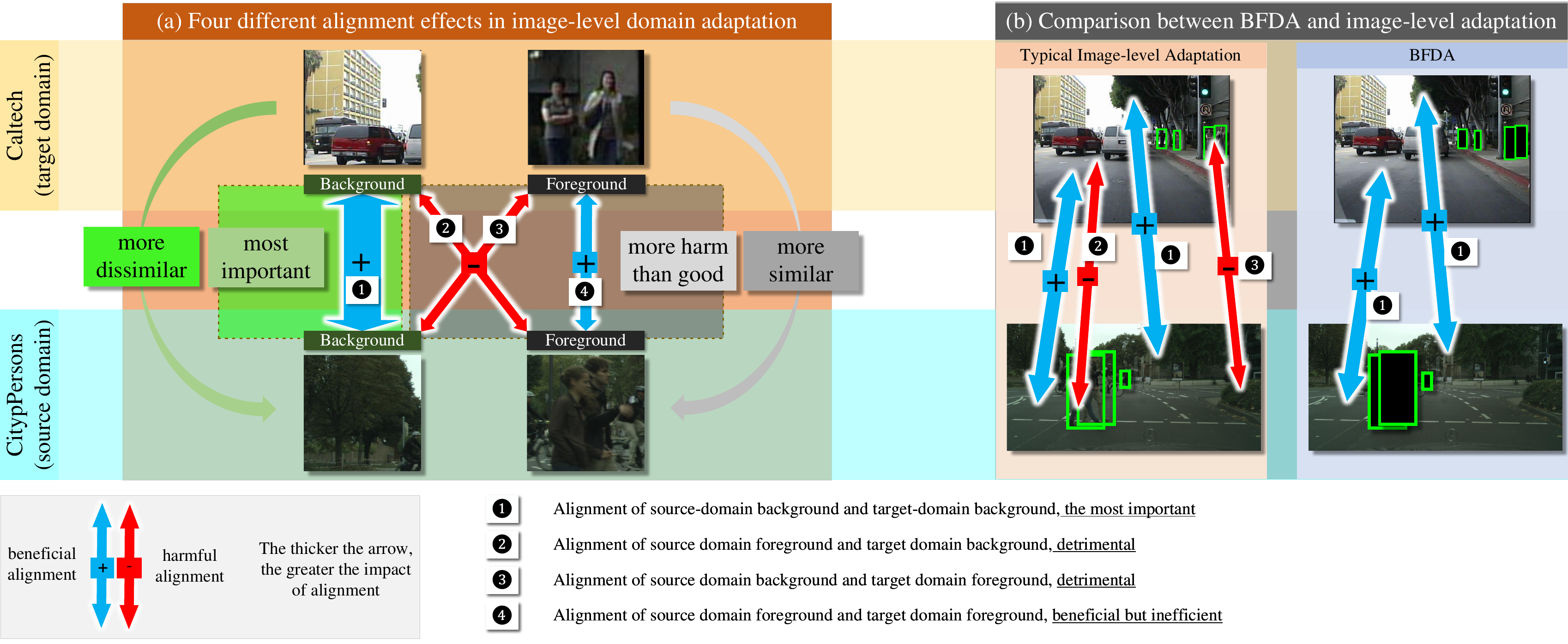}
\caption{The foreground-background feature misalignment issue and the importance of background feature. In Image-level adaptation, the presence of foreground elements can result in simultaneous misalignments (\ding{183},\ding{184},\ding{185}) due to variations in foreground instance positions across different images. We demonstrate that the background alignment \textbf{plays a major role}, and the three alignments (\ding{183},\ding{184},\ding{185}) in the gray box \textbf{cause more harm than good}. Therefore, we prioritize background feature alignment while minimizing interference from foreground elements in image-level cross-domain alignment.} 
  \label{fig:ouyang}
\end{figure*}

\section{Introduction}
\IEEEPARstart{P}{edestrian} detection (PD) is a crucial and longstanding task that has a key role in fields such as autonomous driving~\cite{campmany2016gpu}, pedestrian reidentification~\cite{yan2021anchor}, and video surveillance~\cite{hattori2015learning}. Thanks to convolutional neural networks (CNNs), significant progress has recently been made with PD frameworks. However, current PD methods heavily rely on the consistency assumption between training and test data distributions, which is difficult to guarantee in the real world. As a result, many well-trained PD models~\cite{liu2018learning, lan2018pedestrian, xie2020mask, liu2019high, liu2019efficient} that work well in one environment, such as clear weather conditions, may fail to generalize to other environments, such as dense fog, heavy rain, or lighting variations, resulting in a significant drop in performance.

Researchers have proposed several cross-domain PD methods to address the above problem ~\cite{guo2019domain,liu2016unsupervised,jiao2021san,kieu2020task,guan2019unsupervised,vazquez2012unsupervised,chen2021box}. These methods aim to alleviate the performance degradation in cross-domain scenarios by aligning the source and target domains, either at the image-level or instance-level. For example, DA-Faster-RCNN~\cite{chen2018domain} leverages image-level and instance-level adaptation to improve the detector's domain generalization ability. Additionally, a selective alignment network (SAN)~\cite{jiao2021san} has been proposed to reduce inter instance differences by aligning each subtype of instances. However, these methods are all based on two-stage detectors (\textit{e.g.}, Faster RCNN\cite{ren2015faster}) that are not suitable for practical applications due to their slow inference speed. On the other hand,  one-stage detectors (\textit{e.g.}, YOLOv5\footnote{https://github.com/ultralytics/YOLOv5}) have sufficient speed but lack instance-level proposals, making it difficult to utilize mainstream instance-level feature alignment algorithms. Therefore,  one-stage cross-domain detectors mainly rely on image-level feature alignment.

However, one issue with image-level feature alignment is the misalignment between the background and foreground, as indicated by the red arrows in Figure~\ref{fig:ouyang}, which has not yet been resolved. As a result, one-stage pedestrian detectors suffer a significant decrease in cross-domain accuracy due to imperfect alignment algorithms. In this study, we aim to mitigate the foreground-background misalignment issue\footnote{Note that we believe that instance-level adaptation does not encounter the foreground-background feature misalignment issue because instance-level proposals naturally separate foreground and background for alignment. However,  one-stage detectors lack the necessary conditions for instance-level adaptation and can only perform image-level domain adaptation.} by focusing on the alignment of cross-domain background features and avoiding the involvement of foreground features in cross-domain alignment. Specifically, we investigate the relative importance of the foreground and background in cross-domain tasks and reveal an essential observation that background alignment plays a crucial role in the domain adaptive pedestrian detection task. Our approach is based on the following two findings:
\IEEEpubidadjcol

First, image-level adaptation directly suffers from the foreground-background feature misalignment issue for dense prediction tasks due to variable instance positions in different images. The alignment strategy proves ineffective, as shown in Figure~\ref{fig:process}. As seen, the highest feature response peaks correspond to the foreground regions, \textit{i.e.}, pedestrians. However, since the pedestrian foreground positions vary across images, the same spatial position may be occupied by the pedestrian foreground in one source image and the background in another target image. Consequently, the image-level adaptation process may align the foreground with the background and vice versa, leading to erroneous results (as demonstrated in Figure~\ref{fig:ouyang}). The foreground-background feature misalignment issue is the core issue in cross-domain PD that we aim to solve.

Second, ensuring background feature consistency across domains is crucial for successful cross-domain pedestrian detection (PD). To illustrate this point, we conducted preliminary studies on two popular detectors: the  one-stage detector YOLOv5 and the two-stage detector Faster RCNN. We discovered that changes in background regions had a more significant impact on accuracy than changes in foreground regions. This finding highlights current pedestrian detectors' sensitivity to background variations. We also noticed that short-range background changes had a greater impact on detection accuracy than long-range background changes, which suggests that contextual background information near pedestrians is more critical for their position predictions. Based on this insight, we focused on background alignment to address the foreground-background feature misalignment issue.

Inspired by the above two insights, we are motivated to rethink the cross-domain PD pipeline to alleviate the negative impact of the foreground-background feature misalignment issue on existing  one-stage detectors. Our method focuses on the background features' cross-domain alignment while mitigating the interference of foreground features. Specifically, we first introduce a Background Decoupling Module that takes feature maps from the detection head and decouples background features using a Feature Generation Module inspired by CycleGAN's algorithm~\cite{zhu2017unpaired} to solve the foreground-background feature misalignment issue as mentioned above. Second, we propose a long-short-range domain discriminator that employs a Transformer-CNN-based parallel structure to allocate global and local attention to different background ranges depending on their proximity to pedestrian instances. Comprehensive testing indicates a significant performance improvement with our novel scheme.

Our main contributions can be summarized as follows:
\begin{itemize}
    \item[(1)] 
    Our research reveals the foreground-background feature misalignment issue that  one-stage pedestrian detectors face when performing image-level feature alignment. Additionally, we have discovered that achieving cross-domain PD requires ensuring the interdomain consistency of background features, which has been a critical but underappreciated aspect. To the best of our knowledge, we are the first to propose focusing on background alignment in cross-domain detection.
    \item[(2)] 
    A new background-focused cross-domain PD framework is proposed, consisting of three key modules: the background decoupling module (\textit{BDM}), the feature generation module (\textit{FGM}), and the parallel Transformer-CNN-based long-short-range domain discriminator (\textit{LSD}). This framework can efficiently mitigate the foreground-background feature misalignment issue by decoupling the background feature from the original feature maps to achieve pure background feature alignment.
    \item[(3)]  Experiments on cross-domain PD are conducted using BFDA, and the results indicate that the proposed BFDA is capable of delivering state-of-the-art performance on the  one-stage detector YOLOv5.
\end{itemize}

\section{Related Work} \label{sec:related}

\subsection{Pedestrian Detection.} The rise of deep learning technology has promoted the development of PD research~\cite{liu2018learning,lan2018pedestrian,xie2020mask,liu2019efficient,liu2019high,li2020box,zhang2020attribute}. They can be roughly divided into anchor-based and anchor-free methods. Anchor-based methods detect objects in a given image by classifying and regressing anchor boxes. They can be subdivided into one-stage and two-stage methods, in which the two-stage methods generate proposals and then calculate the confidence score for each proposal. For example, MGAN's~\cite{xie2020mask} attention network emphasizes visible pedestrian areas while adjusting physical characteristics to suppress occluded areas. The one-stage methods directly process the detection, classification, and regression in one step. For example, ALFNet~\cite{liu2018learning} stacks a series of predictors to gradually evolve the default anchor boxes of SSD~\cite{liu2016ssd}. Anchor-free methods such as CSP~\cite{liu2019high} focus on other pedestrian features such as the center and corners. However, these methods face a significant challenge when testing images with widely varying feature distributions. We intend to overcome such a challenge in this work. Before deep learning, some traditional vision works studied how to use the background to solve tracking tasks \cite{wu2013online,zhang2005improving,nguyen2006robust}, but our work differs from these works in both the background exploration method and the aiming task.

\subsection{Cross-domain Object Detection.}
Focusing on the problem of the detector being unable to be generalized to datasets with significant domain gaps, cross-domain object detection technology is proposed. As the pioneer in this field, \cite{chen2018domain} proposed image-level and instance-level adaptation methods and then aligned these features simultaneously. \cite{saito2019strong} designed an adaptive method based on strong-local and weak-global alignment. \cite{he2021partial} deployed an ancillary net parallel to the chief net and formulated an asymmetric tri-way architecture to avoid model collapse in the aligning procedure. \cite{wang2021afan} integrated the intermediate domain image generator and multiscale adversarial feature alignments into a single framework to progressively bridge the domain divergence. \cite{chen2021sequential}  introduced a reinforcement learning-based method to gradually refine source and target instances and alleviate the negative transfer. However, the above methods are based on two-stage detection frameworks, which depend highly on region proposal and the region feature based on ROI pooling. For one-stage detectors, \cite{kim2019self} introduces a weak self-training method to suppress the effects of false-negatives and false-positives and adversarial background score regularization to extract discriminative features for target backgrounds to aid foreground alignment. \cite{zhang2021densely} proposed a semantic enhancement module to strengthen the foreground and multiscale features for cross-domain adaptation. \cite{chen2021i3net} proposed reweighting the image level alignment procedure and matching the pattern of foreground objects guided by the categorical information.  \cite{vs2021mega} addressed the conflict among foreground classes. Unfortunately, these methods are not designed for pedestrians as objects and face the foreground-background feature misalignment issue.\textcolor{black}{~\cite{tang2022source} proposed a three-step method to mine and refine foreground pseudo labels in the target domain. \cite{wu2022unreliability} proposed a generative method based on disentanglement to produce diverse and reliable pedestrian instances for improving the discriminative module's recognition ability and the quality of pseudo labels on the target domain. These two methods ignore the impact of background shift on the pseudo labels.} These cross-domain frameworks may be suboptimal when recognizing pedestrians with diverse appearances. \textcolor{black}{Some methods~\cite{huang2021learning,yang2022learning,yang2021partially} also studied the misalignment, and they mainly focus on establishing the cross-view and cross-modal correspondence between the training samples.} In the method presented in this paper, we discard all foreground classes and only study the background (more complex than a class and can contain arbitrary objects) for the purpose of alignment.


\subsection{Transformer in Vision.} Transformers originate from natural language processing. ViT ~\cite{dosovitskiy2020image} demonstrates that an improved transformer can achieve SOTA results in the image classification task with sufficient data (e.g., ImageNet-22k, JFT-300M). Compared to convolutional neural networks, which typically have limited local attention, the Transformer's global attention mechanism makes it suitable for various computer vision tasks. For instance, the DETR model~\cite{carion2020end} applies a transformer encoder-decoder architecture to object detection with success. To address the storage and computation demands of the transformer, some recent methods~\cite{wu2021cvt, wang2021pyramid, wang2021pvtv2, huang2019ccnet} have reduced the number of parameters used. For example, the CvT model~\cite{wu2021cvt} adopts convolutional token embedding and convolutional projection techniques. \textcolor{black}{~\cite{qu2022distillation} proposed distilling the knowledge from Oracle Queries to address the semantic ambiguity of HOI query problems on the Human-Object Interaction Detection task.}
In our work, we combine the Transformer's global attention capabilities with a short-range convolutional attention module to create the long-short-range domain discriminator. 

\section{Rethinking Cross-domain PD} \label{sec3}

This section begins with an overview of image-level cross-domain adaptation, which is the most significant algorithm for one-stage PD detectors. We also discuss some key observations regarding the theory of the discriminator. Next, we demonstrate the foreground-background feature misalignment issue and highlight the importance of the background in PD tasks through experiments. Finally, we propose a new cross-domain PD paradigm based on theoretical considerations.

\subsection{Typical Image-level Adaptation} \label{sec3.1}
In 2010, ~\cite{ben2010theory} demonstrated the effectiveness of domain discriminators in reducing the difference $d_{\mathcal{H} \Delta \mathcal{H}}\left(\mathcal{D}_{S}, \mathcal{D}_{T}\right)$ in feature distribution $F(I)$ between domains.  The study also showed that domain discriminators can limit the upper bound of the domain generalization error of a classifier model in the target domain.

DA-Faster-RCNN~\cite{chen2018domain} represented the first formulation of the image-level adaptation. Object detection can be regarded as learning the posterior distribution $P(C, B|I)$, where $I$ is the image representation, $B$ are the bounding boxes of objects, and $C \in\{1, \cdots, K\}$ are different classes. The joint distribution of training samples can be expressed as $P(C,B,I)$, with $P_{\mathcal{S}}(C, B, I)$ and $P_{\mathcal{T}}(C, B, I)$ representing the joint distribution of source and target samples, respectively. When domain gaps exist,
$P_{\mathcal{S}}(C, B, I) \neq P_{\mathcal{T}}(C, B, I)$.
The joint distribution can be decomposed into:
\begin{equation}
  P(C, B, I)=P(C, B|I) P(I)
  \label{eq:ty6}.
\end{equation}
According to the covariate shift assumption~\cite{sugiyama2008direct}:
\begin{equation}
  P_{\mathcal{S}}(C, B|I)=P_{\mathcal{T}}(C, B|I)
  \label{eq:ty7}.
\end{equation}
Researchers are committed to using adversarial feature learning methods to train a feature extractor $F$ so that:
\begin{equation}
  P_{\mathcal{S}}(F(I)) \approx P_{\mathcal{T}}(F(I))
  \label{eq:ty8},
\end{equation}
\begin{equation}
  P_{\mathcal{S}}(C, B, F(I)) \approx P_{\mathcal{T}}(C, B, F(I))
  \label{eq:ty9}.
\end{equation}

In fact, DA-Faster-RCNN~\cite{chen2018domain} directly applies the theoretical proof from~\cite{ben2010theory} to the cross-domain object detection task. However, it is worth noting that the theory in ~\cite{ben2010theory} is specifically designed for the cross-domain image classification task. Let us review the key proof steps (Theorem 1 in ~\cite{ben2010theory}). ~\cite{ben2010theory} defines a domain as a pair consisting of a distribution $\mathcal{D}$ on inputs $\mathcal{X}$ and a labeling function $f: \mathcal{X} \rightarrow[0,1]$. They are denoted by $\left\langle\mathcal{D}_{S}, f_{S}\right\rangle$ the source domain and $\left\langle\mathcal{D}_{T}, f_{T}\right\rangle$ the target domain. A hypothesis is a function $h: \mathcal{X} \rightarrow\{0,1\}$. The probability according to the distribution $\mathcal{D}_{S}$ that a hypothesis $h$ disagrees with a labeling function $f$ (which can also be a hypothesis) is defined as $\epsilon_{S}(h, f)=\mathrm{E}_{\mathbf{x} \sim \mathcal{D}_{S}}[|h(\mathbf{x})-f(\mathbf{x})|]$.

\begin{equation}
\footnotesize
\begin{aligned}
\epsilon_{T}(h) &=\epsilon_{T}(h)+\epsilon_{S}(h)-\epsilon_{S}(h)+\epsilon_{S}\left(h, f_{T}\right)-\epsilon_{S}\left(h,_{T}\right) \\
& \leq \epsilon_{S}(h)+\left|\epsilon_{S}\left(h, f_{T}\right)-\epsilon_{S}\left(h, f_{S}\right)\right|+\left|\epsilon_{T}\left(h, f_{T}\right)-\epsilon_{S}\left(h, f_{T}\right)\right| \\
& \leq \epsilon_{S}(h)+\mathrm{E}_{\mathcal{D}_{S}}\left[\left|f_{S}(\mathbf{x})-f_{T}(\mathbf{x})\right|\right]+\left|\epsilon_{T}\left(h, f_{T}\right)-\epsilon_{S}\left(h, f_{T}\right)\right| \\
& \leq \epsilon_{S}(h)+\mathrm{E}_{\mathcal{D}_{S}}\left[\left|f_{S}(\mathbf{x})-f_{T}(\mathbf{x})\right|\right]+\\
& \int\left|\phi_{S}(\mathbf{x})-\phi_{T}(\mathbf{x})\right|\left|h(\mathbf{x})-f_{T}(\mathbf{x})\right| d \mathbf{x} \\
& \leq \epsilon_{S}(h)+\underline{\mathrm{E}_{\mathcal{D}_{S}}\left[\left|f_{S}(\mathbf{x})-f_{T}(\mathbf{x})\right|\right]}+d_{1}\left(\mathcal{D}_{S}, \mathcal{D}_{T}\right)
\end{aligned}
\label{eq:AT}.
\end{equation}

We only need to observe the underlined item, which is considered small in the original text in ~\cite{ben2010theory} and discarded (Page 155): "... and the third is the difference in labeling functions across the two domains, which we expect to be small." Although this assumption may hold for image classification tasks, it does not apply to object detection tasks. In object detection tasks, the underlined item represents the intradomain gap in feature maps and cannot be ignored. Unlike image classification or semantic segmentation tasks, the intradomain gap of feature maps is larger in object detection tasks due to foreground peaks (as shown in Figure \ref{fig:process}) being at different positions. Therefore, foreground feature alignment is not a preferred part of image-level cross-domain.

\subsection{The foreground-background feature misalignment issue} \label{sec3.2}
\begin{figure*}
\centering
\includegraphics[width=1\linewidth]{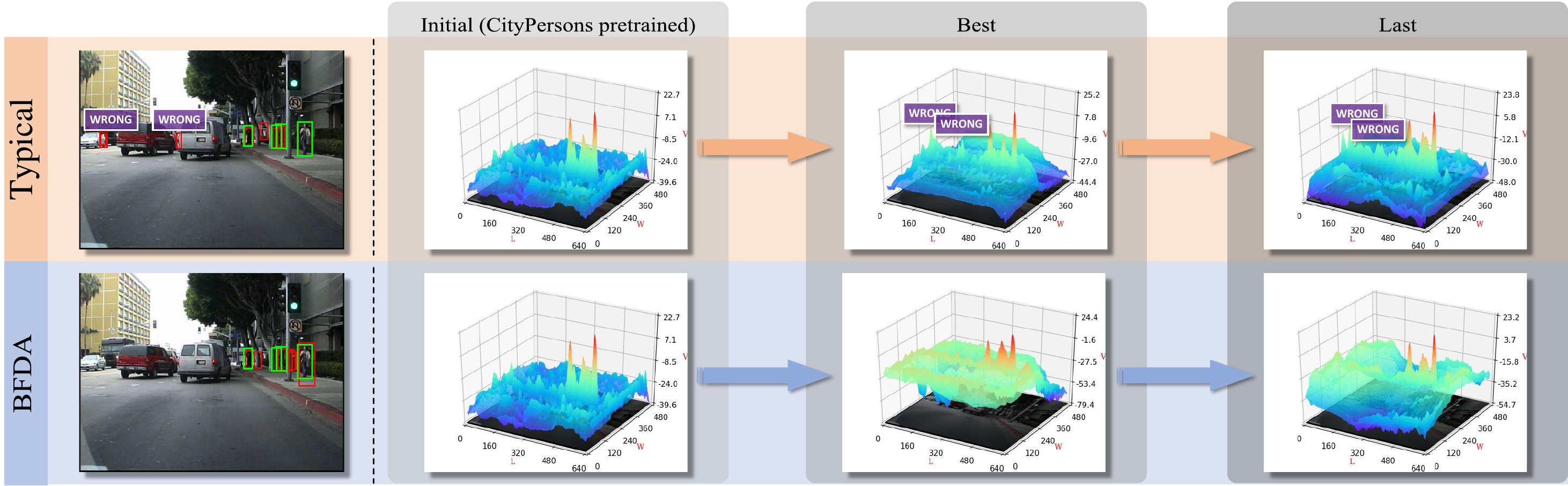}
\caption{Illustration of the feature map evolving process of the typical image-level adaptation (first row) and BFDA (second row) (\textit{CityPersons$\rightarrow$Caltech}, \textbf{test on Caltech}). The left part denotes the detection results (green and red boxes denote the ground truths and predictions, respectively). We study the feature map with the largest resolution after the first convolution layer in the detection head of YOLOv5. Initial, Best, and Last represent the beginning of cross-domain training (pretrained), the epoch with the best detection results, and the last epoch, respectively. In the 3D feature map visualization, two horizontal axes represent the spatial dimensions, while the vertical axis represents the channel dimension (sum along the channel). It can be seen that, due to the misalignment between foreground instances and background regions, \textbf{wrong peaks} gradually appear on the feature map during training, resulting in false detections (purple boxes) when typical image-level adaptation is used. In fact, this is because the background features of some images are aligned with the foreground features of other images, so the detector's ability to distinguish the foreground and background of the image decreases. We call it \textbf{the foreground-background feature misalignment issue}. Our method alleviates this problem by focusing on the background feature alignment between domains and reducing the interference of the foreground in image-level cross-domain feature alignment.}
  \label{fig:process}
\end{figure*}

However, the abovementioned image-level domain adaptation methods based on adversarial feature learning inevitably face the foreground-background feature misalignment issue. To illustrate this issue more clearly, we visualize the change in feature maps throughout the process of image-level cross-domain alignment (as shown in Figure \ref{fig:process}).  

The alignments can bring nonnegligible hazards in dense prediction tasks such as object detection, especially PD. That is, the \textbf{foreground-background feature misalignment issue}. The feature map's most prominent parts (peaks) represent possible pedestrian instances, which are also of the most concern during the feature alignment process. However, the location information of pedestrian instances in different images may be quite different. In this case, typical image-level adaptation results in many foreground regions of some images incorrectly aligning with the background regions of other images. As a result, the features in background regions are also inevitably accounted for in the alignment process, leading to fake peaks (wrong boxes in Figure \ref{fig:process}), which leads to false detections.

\begin{table*}
\begin{center}
\caption{Ablation study of feature change in different regions of images. We use YOLOv5 and Faster RCNN to conduct experiments on the CityPersons~\cite{Shanshan2017CVPR} and BDD10K~\cite{bdd100k} datasets. BDD10K does not provide occlusion labels, so partial and heavy (section \ref{sec5.2}) cannot be reported. Obviously, changes in background features have a greater impact on detection accuracy than the foreground feature change.
($*$ The result in the seventh-to-last row is correct.)}
\label{table:dif_area}
\renewcommand\arraystretch{1.2}
\begin{tabular}{c|c|c|c|c|cccc|c}
\toprule
\multirow{2}{*}{Method} & \multirow{2}{*}{Dataset} & \multirow{2}{*}{Foreground} & \multirow{2}{*}{\begin{tabular}[c]{@{}c@{}}Inner-bounding-box \\ background\end{tabular}} & \multirow{2}{*}{\begin{tabular}[c]{@{}c@{}}Outer-bounding-box \\ background\end{tabular}} & \multicolumn{4}{c|}{$M R^{-2} (\%)$ ↓} & \multirow{2}{*}{$\operatorname{AP}_{50}(\%)$ ↑} \\ \cline{6-9}
 &  &  &  &  & \multicolumn{1}{c|}{\textbf{reasonable}} & \multicolumn{1}{c|}{bare} & \multicolumn{1}{c|}{partial} & heavy &  \\ \hline\hline
YOLOv5 & CityPersons & \checkmark & \checkmark & \checkmark & \multicolumn{1}{c|}{10.45} & \multicolumn{1}{c|}{7.38} & \multicolumn{1}{c|}{8.63} & 40.28 & 83.5 \\ \hline
YOLOv5 & CityPersons & \checkmark & \checkmark &  & \multicolumn{1}{c|}{18.74} & \multicolumn{1}{c|}{11.02} & \multicolumn{1}{c|}{19.33} & 46.22 & 65.1(-18.4) \\ \hline
YOLOv5 & CityPersons & \checkmark &  & \checkmark & \multicolumn{1}{c|}{34.56} & \multicolumn{1}{c|}{26.53} & \multicolumn{1}{c|}{37.76} & 72.99 & 56.0(-27.5) \\ \hline
YOLOv5 & CityPersons &  & \checkmark & \checkmark & \multicolumn{1}{c|}{38.07} & \multicolumn{1}{c|}{31.04} & \multicolumn{1}{c|}{40.78} & 71.16 & 58.9(-24.6) \\ \hline
YOLOv5 & CityPersons & \checkmark &  &  & \multicolumn{1}{c|}{47.36} & \multicolumn{1}{c|}{41.33} & \multicolumn{1}{c|}{46.96} & 74.43 & 47.5(-36.0) \\ \hline\hline
YOLOv5 & BDD10K & \checkmark & \checkmark & \checkmark & \multicolumn{1}{c|}{13.46} & \multicolumn{1}{c|}{13.46} & \multicolumn{1}{c|}{-} & - & 73.8 \\ \hline
YOLOv5 & BDD10K & \checkmark & \checkmark &  & \multicolumn{1}{c|}{50.20} & \multicolumn{1}{c|}{50.20} & \multicolumn{1}{c|}{-} & - & 25.5(-48.3)  \\ \hline
YOLOv5 & BDD10K & \checkmark &  & \checkmark & \multicolumn{1}{c|}{18.02} & \multicolumn{1}{c|}{18.02} & \multicolumn{1}{c|}{-} & - & 68.8(-5.0) \\ \hline
YOLOv5 & BDD10K &  & \checkmark & \checkmark & \multicolumn{1}{c|}{11.28} & \multicolumn{1}{c|}{11.28} & \multicolumn{1}{c|}{-} & - & 82.5(+8.7)$^*$ \\ \hline
YOLOv5 & BDD10K & \checkmark &  &  & \multicolumn{1}{c|}{38.55} & \multicolumn{1}{c|}{38.55} & \multicolumn{1}{c|}{-} & - & 39.7(-34.1) \\ \hline\hline
Faster RCNN & CityPersons & \checkmark & \checkmark & \checkmark & \multicolumn{1}{c|}{19.65} & \multicolumn{1}{c|}{10.84} & \multicolumn{1}{c|}{21.33} & 85.10 & 68.3 \\ \hline
Faster RCNN & CityPersons & \checkmark & \checkmark &  & \multicolumn{1}{c|}{69.13} & \multicolumn{1}{c|}{63.83} & \multicolumn{1}{c|}{67.00} & 93.25 & 20.5(-47.8) \\ \hline
Faster RCNN & CityPersons & \checkmark &  & \checkmark & \multicolumn{1}{c|}{45.53} & \multicolumn{1}{c|}{34.98} & \multicolumn{1}{c|}{51.33} & 87.96 & 35.4(-32.9) \\ \hline
Faster RCNN & CityPersons &  & \checkmark & \checkmark & \multicolumn{1}{c|}{56.29} & \multicolumn{1}{c|}{46.43} & \multicolumn{1}{c|}{62.49} & 94.90 & 35.5(-32.8) \\ \hline
Faster RCNN & CityPersons & \checkmark &  &  & \multicolumn{1}{c|}{52.68} & \multicolumn{1}{c|}{41.35} & \multicolumn{1}{c|}{54.22} & 91.16 & 31.9(-36.4) \\  
\bottomrule
\end{tabular}
\end{center}
\end{table*}

\subsection{The importance of background in pedestrian detection} \label{sec3.3}

\begin{figure}[t]
  \centering
  \includegraphics[width=1\columnwidth]{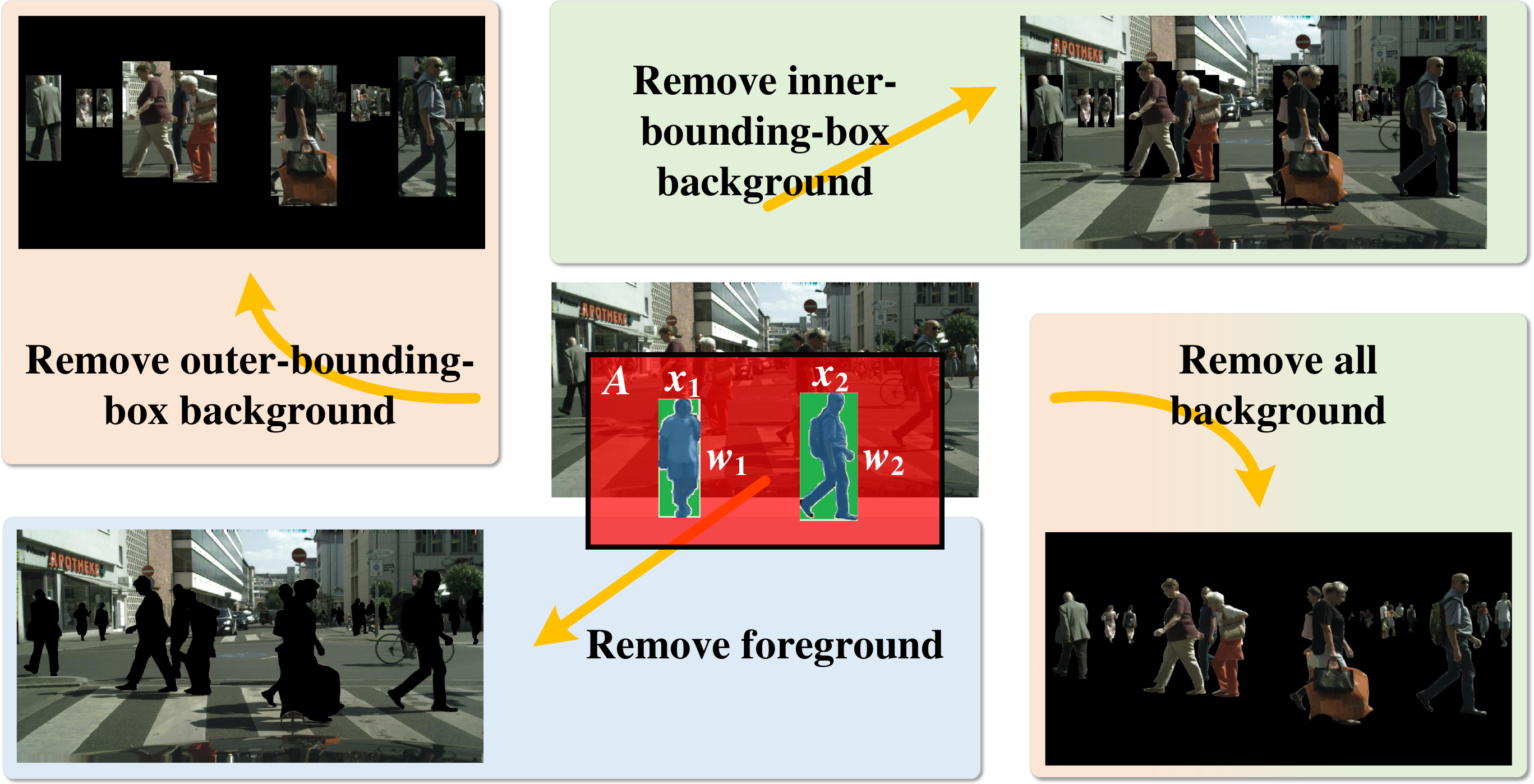}

  \caption{Feature change in different regions of background. We blacken different regions of the image (represents that the features in this region have changed). In these cases, we can study the dependence of mainstream detectors on each part of the feature.
  }
  \label{fig:background}
\end{figure}
\begin{figure}[t]
  \centering
  \includegraphics[width=1\columnwidth]{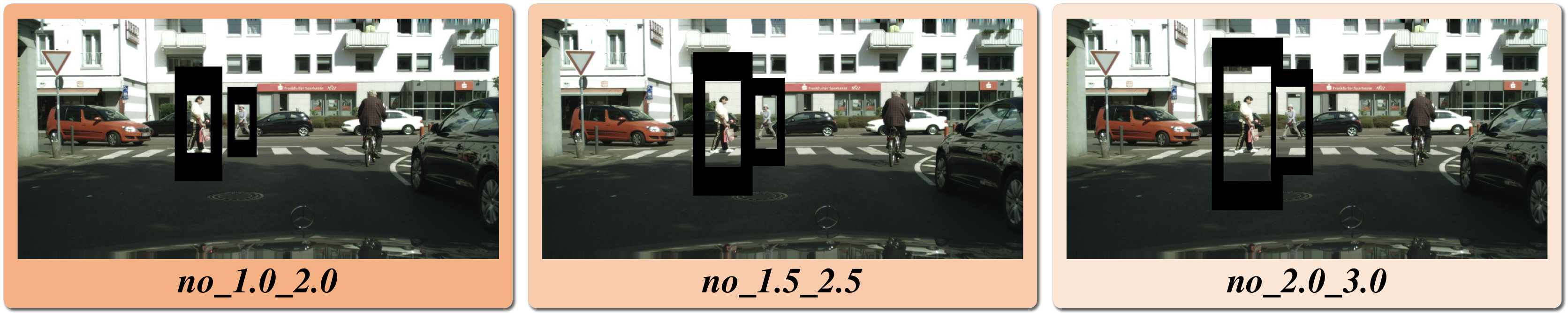}

  \caption{
 Feature change in different ranges of background. $no \_ x \_ y$ represents the images whose background features (from $x$ times to $y$ times the size of the bounding boxes) have changed.
  }
  \label{fig:LS_range}
\end{figure}

\begin{table*}
\begin{center}
\caption{Ablation study of feature change in different ranges of backgrounds. Changes in the short-range background feature significantly impact detection accuracy more than the long-range background feature change.}
\label{table:dif_range} 
\renewcommand\arraystretch{1.2}
\setlength{\tabcolsep}{5.2mm}{
\begin{tabular}{c|c|c|cccc|c}
\toprule
\multirow{2}{*}{Method} & \multirow{2}{*}{Dataset} & \multirow{2}{*}{Range} & \multicolumn{4}{c|}{$M R^{-2} (\%)$ ↓} & \multirow{2}{*}{$\operatorname{AP}_{50}(\%)$ ↑} \\ \cline{4-7}
 &  &  & \multicolumn{1}{c|}{\textbf{reasonable}} & \multicolumn{1}{c|}{bare} & \multicolumn{1}{c|}{partial} & heavy &  \\ \hline\hline
YOLOv5 & CityPersons & \textit{original image} & \multicolumn{1}{c|}{10.45} & \multicolumn{1}{c|}{7.38} & \multicolumn{1}{c|}{8.63} & 40.28 & 83.5 \\\hline
YOLOv5 & CityPersons & \textit{no\_1.0\_2.0} & \multicolumn{1}{c|}{25.30} & \multicolumn{1}{c|}{18.96} & \multicolumn{1}{c|}{26.86} & 55.80 & 58.9(-24.6) \\\hline
YOLOv5 & CityPersons & \textit{no\_1.5\_2.5} & \multicolumn{1}{c|}{12.50} & \multicolumn{1}{c|}{7.77} & \multicolumn{1}{c|}{11.54} & 41.96 & 75.6(-7.9) \\\hline
YOLOv5 & CityPersons & \textit{no\_2.0\_3.0} & \multicolumn{1}{c|}{10.21} & \multicolumn{1}{c|}{6.88} & \multicolumn{1}{c|}{8.22} & 39.24 & 81.0(-2.5) \\ \hline\hline
YOLOv5 & BDD10K & \textit{original image} & \multicolumn{1}{c|}{13.46} & \multicolumn{1}{c|}{13.46} & \multicolumn{1}{c|}{-} & - & 73.8 \\\hline
YOLOv5 & BDD10K & \textit{no\_1.0\_2.0} & \multicolumn{1}{c|}{46.92} & \multicolumn{1}{c|}{46.92} & \multicolumn{1}{c|}{-} & - & 27.2(-46.6) \\\hline
YOLOv5 & BDD10K & \textit{no\_1.5\_2.5} & \multicolumn{1}{c|}{20.64} & \multicolumn{1}{c|}{20.64} & \multicolumn{1}{c|}{-} & - & 49.2(-24.6) \\\hline
YOLOv5 & BDD10K & \textit{no\_2.0\_3.0} & \multicolumn{1}{c|}{16.31} & \multicolumn{1}{c|}{16.31} & \multicolumn{1}{c|}{-} & - & 58.1(-15.7) \\ \hline\hline
Faster RCNN & CityPersons & \textit{original image} & \multicolumn{1}{c|}{19.65} & \multicolumn{1}{c|}{10.84} & \multicolumn{1}{c|}{21.33} & 85.10 & 68.3 \\\hline
Faster RCNN & CityPersons & \textit{no\_1.0\_2.0} & \multicolumn{1}{c|}{76.15} & \multicolumn{1}{c|}{72.57} & \multicolumn{1}{c|}{76.57} & 94.23 & 16.6(-51.7) \\\hline
Faster RCNN & CityPersons & \textit{no\_1.5\_2.5} & \multicolumn{1}{c|}{53.86} & \multicolumn{1}{c|}{46.93} & \multicolumn{1}{c|}{54.77} & 89.41 & 35.6(-32.7) \\\hline
Faster RCNN & CityPersons & \textit{no\_2.0\_3.0} & \multicolumn{1}{c|}{37.65} & \multicolumn{1}{c|}{28.92} & \multicolumn{1}{c|}{39.54} & 86.94 & 47.6(-20.7)\\
\bottomrule

\end{tabular}}
\end{center}
\end{table*}

Considering the foreground-background feature misalignment issue, we investigate whether cross-domain pedestrian detection can be achieved by aligning \textbf{only} the background, which has never been considered before. To provide persuasive results, we conduct experiments on the CityPersons and BDD10k datasets using the mainstream detector YOLOv5. Additionally, we use the mainstream two-stage detector Faster RCNN to investigate whether the importance of the background is specific to one-stage detectors. The general evaluation metrics $M R^{-2}(\%)$ for pedestrian detection and $\mathrm{AP}_{50}(\%)$ for object detection are used for performance comparison. \textbf{A smaller (↓) $M R^{-2}(\%)$ and larger (↑) $\mathrm{AP}_{50}(\%)$ indicate that the method is better.} The PD task uses $M R^{-2}$ as the standard evaluation metric, and we show $\mathrm{AP}_{50}(\%)$ here to make our findings more convincing.

First, we examine the feature change impact on the accuracy of PD in three regions (outer-bounding-box background, inner-bounding-box background, and foreground), as illustrated in Figure~\ref{fig:background} and summarized in Table~\ref{table:dif_area}. Our experiments reveal that changes in background features significantly affect the accuracy of pedestrian detectors. Thus, during the image-level feature alignment process, it is crucial to prioritize background feature alignment over foreground feature alignment.

Second, we explore how different background regions impact the detection results, considering varying ranges to the foreground instances. This analysis is presented in Figure \ref{fig:LS_range} and Table \ref{table:dif_range}. Our findings reveal that distinct background regions can produce diverse effects on the results. Mainstream detectors are more sensitive to feature change in background regions close to the instance (short-range).

\subsection{New Paradigm: Background-focused Feature Alignment} \label{sec3.4}
Motivated by the above studies, we propose a new paradigm named background-focused distribution alignment. Given an input image, $\boldsymbol{A}$ denotes the outer-bounding-box background (the \textcolor{red}{\textbf{red part}} in Figure \ref{fig:background}). Meanwhile, $\left\{\boldsymbol{x}_{i}\right\}$ and $\left\{\boldsymbol{w}_{i}\right\}, i=1, \ldots, n$ denote the inner-bounding-box background (the \textcolor{green}{\textbf{green part}} in Figure \ref{fig:background}) and pixel-level foreground (the \textcolor{blue}{\textbf{blue part}} in Figure \ref{fig:background}), respectively. $\boldsymbol{F}_{S}$ refers to the detector trained only on the source domain, and $K_S,K_T$ represent the detection performance on the source and target domain test datasets, respectively. We assume that $\left\{\boldsymbol{w}_{k}, \boldsymbol{w}_{q}\right\}, k \neq q$ are independent, and $\left\{\boldsymbol{x}_{k}, \boldsymbol{x}_{q}\right\}, k \neq q$ are independent, which means that (\textit{P} stands for feature probability distribution):
 $
 P\left(\boldsymbol{w_{S}}\right) = \prod_{i=1}^{n_{S}} P\left(\boldsymbol{w}_{Si}\right),
$
 $
 P\left(\boldsymbol{w_{T}}\right) = \prod_{i=1}^{n_{T}} P\left(\boldsymbol{w}_{Ti}\right), 
$
$
 P\left(\boldsymbol{x_{S}}\right) = \prod_{i=1}^{n_{S}} P\left(\boldsymbol{x}_{Si}\right), 
$
$
 P\left(\boldsymbol{x_{T}}\right) = \prod_{i=1}^{n_{T}} P\left(\boldsymbol{x}_{Ti}\right). 
$

Additionally, we have Equations \ref{eq:5} and \ref{eq:6}. $\boldsymbol{F}_{S}$ refers to the detector trained only on the source domain. In simple terms, the detection accuracy is related to the detector and its three inputs $P\left(\boldsymbol{w_{S}}\right), P\left(\boldsymbol{x_{S}}\right), P\left(\boldsymbol{A}_{S}\right)$:
 \begin{equation}
K_{S} \propto \boldsymbol{F}_{S}\left(P\left(\boldsymbol{w_{S}}\right), P\left(\boldsymbol{x_{S}}\right), P\left(\boldsymbol{A}_{S}\right)\right)
  \label{eq:5},
\end{equation}
\begin{equation}
K_{T} \propto \boldsymbol{F}_{S}\left(P\left(\boldsymbol{w_{T}}\right), P\left(\boldsymbol{x_{T}}\right), P\left(\boldsymbol{A}_{T}\right)\right)
  \label{eq:6}.
\end{equation}
 
 We derive the following:
\begin{equation}
K_{S} \propto \boldsymbol{F}_{S}\left(\prod_{i=1}^{n_{S}} P\left(\boldsymbol{w}_{Si}\right), \prod_{i=1}^{n_{S}} P\left(\boldsymbol{x}_{Si}\right), P\left(\boldsymbol{A}_{S}\right)\right)
  \label{eq:new1},
\end{equation} 

\begin{equation}
K_{T} \propto \boldsymbol{F}_{S}\left(\prod_{j=1}^{n_{T}} P\left(\boldsymbol{w}_{Tj}\right), \prod_{j=1}^{n_{T}} P\left(\boldsymbol{x}_{Tj}\right), P\left(\boldsymbol{A}_{T}\right)\right)
  \label{eq:new2}.
\end{equation} 

Using the total differential equation ($w=w(x,y,z) \Longrightarrow  \Delta w=\frac{\partial w}{\partial x} \Delta x+\frac{\partial w}{\partial y} \Delta y+\frac{\partial w}{\partial z} \Delta z$):
\begin{equation}
\begin{split}
\Delta K \propto C_{1} \cdot \frac{\partial \boldsymbol{F}^{e}_{S}}{\partial(P(\boldsymbol{w}))} \cdot \Delta P(\boldsymbol{w}) + C_{2} \cdot \frac{\partial \boldsymbol{F}^{e}_{S}}{\partial(P(\boldsymbol{x}))} \cdot \\ \Delta P(\boldsymbol{x})+ C_{3} \cdot \frac{\partial \boldsymbol{F}^{e}_{S}}{\partial(P(\boldsymbol{A}))} \cdot \Delta P(\boldsymbol{A})
\end{split}
\label{eq:derivative},
\end{equation}
where $C_1$, $C_2$, and $C_3$ are coefficients that represent the coefficients due to the partial derivative of the cumulative product. $\boldsymbol{F}^{e}_{S}$ denotes $\boldsymbol{F}_{S}\left\{P\left(\boldsymbol{w}\right), P\left(\boldsymbol{x}\right), P\left(\boldsymbol{A}\right)\right\}$.
$\Delta$ refers to the domain gap, \textit{e.g.}, $\Delta K = K_{S} - K_{T}$, and $\Delta P(\boldsymbol{A})$ is the gap between $P\left(\boldsymbol{A}_{S}\right)$ and  $P\left(\boldsymbol{A}_{T}\right)$.

The experimental results in Table 1 in the original text demonstrate that (although the influence of the feature changes of each region on the detection accuracy is different, it is of the same order of magnitude.):
\begin{equation}
C_{1} \cdot \frac{\partial \boldsymbol{F}^{e}_{S}}{\partial(P(\boldsymbol{w}))} \approx C_{2} \cdot \frac{\partial \boldsymbol{F}^{e}_{S}}{\partial(P(\boldsymbol{x}))} \approx C_{3} \cdot \frac{\partial \boldsymbol{F}^{e}_{S}}{\partial(P(\boldsymbol{A}))}.
\label{eq:11}
\end{equation}
At the same time, for the cross-domain pedestrian detection task, the interdomain difference of the background is much larger than the interdomain difference of the foreground. In fact, the reason why we feel that the two images have domain differences is mainly because of the domain differences in the background (such as fog or no fog background). The domain difference for pedestrian foreground is not large: 
\begin{equation}
\Delta P(\boldsymbol{w})<\Delta P(\boldsymbol{x}) \ll \Delta P(\boldsymbol{A}).
\label{eq:12}
\end{equation}
In this case, by Equation \ref{eq:derivative}, Equation \ref{eq:11} and Equation \ref{eq:12}, we can obtain:
\begin{equation}
\Delta K \propto O\left(\frac{\partial \boldsymbol{F}^{e}_{S}}{\partial(P(\boldsymbol{A}))} \cdot \Delta P(\boldsymbol{A})\right).
\label{eq:13}
\end{equation}

Therefore, we can make $\frac{\partial \boldsymbol{F}^{e}_{S}}{\partial(P(\boldsymbol{A}))} \rightarrow 0$ through background-focused distribution alignment, greatly reducing $\Delta K = K_{S} - K_{T} (\Delta K \rightarrow 0) $.

Equation \ref{eq:13} msignifies that the inconsistency of background features in cross-domain detection has the greatest impact on accuracy, so we can perform image-level cross-domain feature alignment by focusing only on the background. The background-focused feature alignment can not only play the same cross-domain adaptation role as the original image-level cross-domain feature alignment but also effectively enable the foreground-background feature misalignment issue to be avoided.

\section{Methodology} \label{sec4}
\begin{figure*}
  \centering
  \includegraphics[width=1\linewidth]{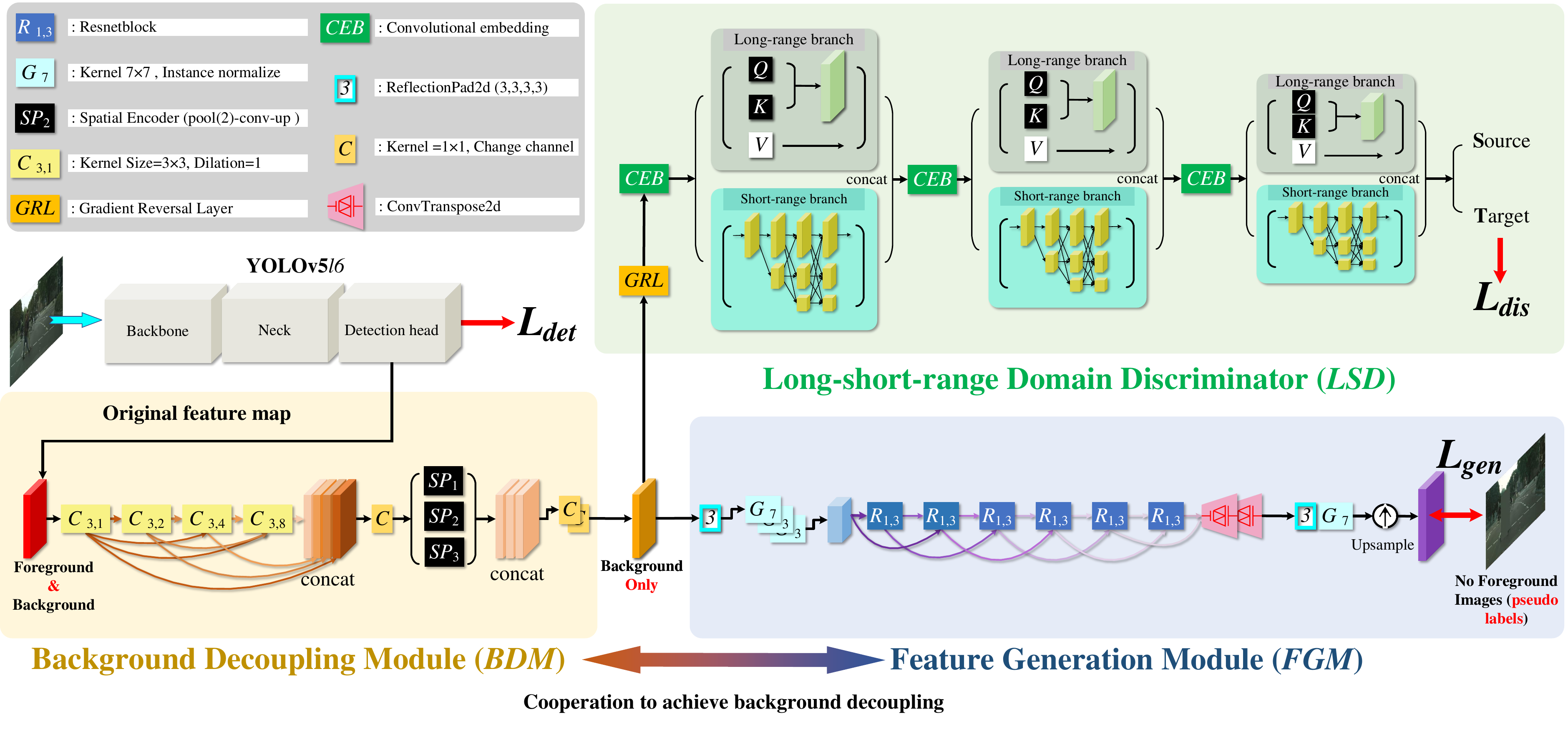}
  \caption{The overall pipeline of the proposed framework consists of four main parts: a \textbf{YOLOv5}\emph{l6} Detector, a Background Decoupling Module extracting background information from the original feature map, a Feature Generation Module generating images containing only the background regions (using \textbf{pseudo labels} to remove the foreground regions of the original images) from the feature map generated by the \textit{BDM}, and a long-short-range domain discriminator which uses the feature map generated by the \textit{BDM} for background-focused distribution alignment. \textit{BDM}, \textit{LSD}, \textit{FGM} are only used during training and will not exist during testing. The function of \textit{BDM} relies on the help of \textit{FGM} to complete.}
  \label{fig:framework}
\end{figure*}

To solve the above issue, we propose three main modules: a background decoupling module (\textit{BDM}), a feature generation module (\textit{FGM}), and a long-short-range domain discriminator (\textit{LSD}), as shown in Figure \ref{fig:framework}. This section introduces these modules from two levels: background feature decoupling and long-short-range attention discriminator.

\subsection{Background Feature Decoupling} \label{sec4.1}

Section \ref{sec3.2} presents the foreground-background feature misalignment issue, and Section \ref{sec3.3} reveals the importance of background in PD. Therefore, our proposed framework decouples the background features from the original feature maps\footnote{The original feature map we studied is extracted from the first layer of convolution in the detection head of YOLOv5} and only aligns the background features between domains. It can perfectly solve the foreground-background feature misalignment issue because no foreground features interfere with the alignment. Our framework mainly decouples background features by the Background Decoupling Module and the Feature Generation Module.

Visual analysis of the original feature maps (Sec. \ref{sec_v} and Figure \ref{fig:fig7}) shows that background and foreground features are fully mixed in the original feature maps. \textbf{It is not possible to completely decouple the background features solely based on spatial position information. } Therefore, the background feature decoupling algorithm we designed is necessary.

First, the Background Decoupling Module specifically targets the decoupling of background features from the subfeature map with the highest resolution in the original feature map\footnote{the original feature map of \textbf{YOLOv5}\emph{l6} (an excellent model in the YOLOv5 series) has four subfeature maps with different resolutions}. This particular subfeature map contains abundant spatial information that effectively characterizes background features while containing limited pedestrian semantic information, making it ideal for background feature decoupling. Despite the insufficient semantic information in the feature map, the \textit{BDM}'s multilevel spatial encoder is capable of analyzing the background's semantic information effectively.

Second, the Feature Generation Module helps the Background Decoupling Module decouple background features (both optimized by the same loss function $L_{gen}$). This module consists of a ResNet-based~\cite{he2016deep} feature encoder and a transposed convolution module, with the aim of reconstructing images containing only the background from the feature map generated by the Background Decoupling Module. Pseudo labels, obtained from the pedestrian detection results of the previous epoch, are used as the ground truth to remove potential foreground regions from the input image. These pseudo labels consist of prediction boxes with confidence scores greater than 0.01, and the corresponding regions in the original image are filled with the \textbf{average pixel value}. We use Manhattan distance to measure the loss:
\begin{equation}
L_{g e n}=\frac{1}{H W}\sum_{i=1}^{H} \sum_{j=1}^{W}\left|I_{i j}-I_{i j}^{\prime}\right|
  \label{eq:lg},
\end{equation}
where the length and width of the image are $H$ and $W$, respectively, and the pixel value at point $(i,j)$ of the ground truth (using pseudo labels) and restored image (by the Feature Generation Module) is expressed as $I_{i j}$ and $I_{i j}^{\prime}$, respectively.

\subsection{Long-short-range: Dual-branch Discriminator} \label{sec4.2}

Figure \ref{fig:LS_range} and Table \ref{table:dif_range} demonstrate that the PD detector exhibits higher sensitivity to short-range background feature changes than to long-range background feature changes.  To effectively analyze both local and global background features, we propose a dual-branch structure comprising a Transformer-CNN-based long-short-range discriminator.

First, to capture global spatial and semantic information from the diverse range of backgrounds in the feature map input to the discriminator, a long-range attention module is crucial. One natural choice for encoding long-distance information is the self-attention-based transformer, which has been shown to be effective in modeling long-range dependencies. Inspired by CvT~\cite{wu2021cvt}, we incorporate a convolutional embedding-based transformer as a long-range branch in our discriminator.

Furthermore, solely focusing on long-range background information is inadequate. In our task, the discriminator must also encode short-range background features and analyze spatial and semantic information, which are crucial. While convolutional networks naturally capture local information, we have designed a short-range attention branch inspired by the multilevel structure of HRNet~\cite{sun2019deep} to address this need. This enables our discriminator to effectively attend to both local and global features for improved performance.

$D_i$ is the domain label of the $i$-th image, and $D_i=0 (1)$ means that the $i$-th image comes from the source (target) domain. $p_i$ is the probability of determining the $i$-th image belonging to the target domain (1) by the discriminator. The discriminator cross-entropy loss can be expressed as:
\begin{equation}
L_{dis}=-\sum_{i}\left[D_{i} \log \left(p_{i}\right)+\left(1-D_{i}\right) \log \left(1-p_{i}\right)\right]
  \label{eq:ld}.
\end{equation}
Adversarial feature alignment requires training the discriminator network to minimize $L_{dis}$ while training the base detector to maximize $L_{dis}$. A gradient reversal layer (GRL) module can help implement this algorithm (Figure \ref{fig:framework}).

The total loss can be expressed as:
\begin{equation}
L=\alpha \cdot L_{d e t}+\beta \cdot L_{g e n}+\gamma \cdot L_{d i s}
  \label{eq:tl},
\end{equation}
where $L_{d e t}$ is the loss when YOLOv5 uses source domain images and annotations for training, and $\alpha$, $\beta$, $\gamma$ are trade-off parameters to balance these losses.

\section{Experiments} \label{sec5}

This section first presents an evaluation of  our proposed BFDA for cross-domain PD tasks in two different scenarios: scene adaptation and weather adaptation. Furthermore, we provide empirical analysis from two perspectives: an ablation study to elucidate the role of BFDA submodules and the generalization performance of BFDA on common object detection tasks. In addition, we present the results and findings of several additional evaluations, including tests on SSD detectors and improvements to the short-range attention mechanism. 

\subsection{Datasets}
\label{sec5.1}

\noindent\textbf{Caltech~\cite{dollar2011pedestrian}} has 42,782 training images and 4024 test images with a resolution of 640$\times$480 pixels. We use the new annotations provided by~\cite{zhang2016far} for experiments. This is one of the most commonly used datasets for pedestrian detection.

\noindent\textbf{CityPersons~\cite{Shanshan2017CVPR}} is built from Cityscapes~\cite{Cordts2016Cityscapes}, which has approximately 2975, 500, and 1525 images for training, validation, and testing, respectively (researchers often use the Cityscapes validation dataset for testing, as we do), with a resolution of  2048$\times$1024 pixels. 

\noindent\textbf{FoggyCityscapes~\cite{sakaridis2018semantic}} is also built from Cityscapes and contains three levels of foggy images. The thickest set was used.

\noindent\textbf{BDD10K~\cite{bdd100k}} is an auxiliary dataset of the BDD100K dataset, which contains 7000, 2000, and 1000 images for training, validation, and testing, respectively, with a resolution of  1280$\times$720 pixels.

Note that Caltech and BDD100K do not include segmentation level annotations, so we can only use BDD10K to generate the dataset we need in Sec. \ref{sec3}.

\subsection{Experiment settings}
\label{sec5.2}

\textbf{Evaluation Settings.}
We utilize two cross-domain settings: (i) \emph{CityPersons$\rightarrow$Caltech}: scene adaptation, where the source domain is CityPersons (Cityscapes), and the target domain is Caltech. (ii) \emph{CityPersons$\rightarrow$Foggy Cityscapes}: weather adaptation, where the source domain is CityPersons (Cityscapes), and the target domain is Foggy Cityscapes.

\textbf{Metrics.}
(i) We utilize the standard log average miss rate over false-positive per image (FPPI) in the range of $\left[10^{-2}, 10^{0}\right]$, dubbed by $M R^{ -2}$ . A lower $M R^{-2}$ indicates better performance. To gain a deeper understanding of the model's performance under different occlusion conditions, we further divide the test set into four parts according to the degree of occlusion (reasonable, bare, partial, and heavy) and report the results separately.
(ii) We also use the general object detection evaluation metric $\operatorname{AP}_{50}$ in section \ref{sec3}. 
PD task uses $M R^{-2}$ (not $\operatorname{AP}_{50}$) as the standard evaluation metric (Therefore, we only report $M R^{-2}$ in Section \ref{sec5}).

\textbf{Implementation Details.}
We follow the standard protocols of UDA, where all samples in the source are labeled while those in the target are unlabeled. BFDA employs the excellent one-stage detector \textbf{YOLOv5}\emph{l6} as the base detector. Since many  SOTA UDA methods designed for two-stage detectors are inapplicable tn  one-stage detectors (\textit{e.g.}, YOLOv5), the Faster RCNN backbone is used in Table \ref{table:all_class} for a fair comparison. The input images maintain their original resolution, but their feature maps are resized to 224$\times$224 before being fed into the discriminator.
When performing cross-domain adaptation, we first initialize the model with pretrained weights. The initial learning rate is $10^{-3}$, which is reduced to $2\times10^{-4}$ by cosine annealing, and the learning rates of the other three modules are $10^{-4}$. Additionally, we do not use the mosaic trick. We use NVIDIA Tesla V100 to test the FPS of the frameworks.

\subsection{Comparison Results}
\label{sec5.3}

\textbf{scene adaptation.}
Scenes captured by different devices or setups often exhibit domain shifts. To evaluate the effectiveness of our proposed framework for scene adaptation, we conduct experiments using CityPersons as the source domain and Caltech as the target domain.

Table~\ref{table:scen} compares our BFDA with current SOTA cross-domain PD models. Although the SOTA cross-domain PD model SAN~\cite{jiao2021san} achieves impressive performance, our overall framework (BFDA) surpasses it by a significant margin, almost reaching the accuracy of \textbf{Oracle} (trained on the labeled target domain dataset). This demonstrates the high effectiveness of our framework in the scene adaptation task. One of the main reasons for this success is that the domain gap of foreground (pedestrian) features in the scene adaptation task is relatively small, while the domain difference of background features plays a major role in performance degradation. Therefore, our background-focused distribution alignment method effectively mitigates the misalignment issue. Furthermore, by utilizing an efficient one-stage base detector, our method achieves a much higher frames per second (FPS) compared to those based on Faster R-CNN, making it suitable for real-time applications such as autonomous driving. 

\textbf{weather adaptation.}
The impact of weather changes is an unavoidable factor that can result in domain gaps in real-world applications, leading to a significant degradation in model performance, as indicated in~\cite{chen2018domain,wang2021domain}. In our investigation of the effectiveness of BFDA, we utilize CityPersons as the source domain and Foggy Cityscapes as the target domain.

Table~\ref{table:weat} presents the results on weather adaptation. The performance of the existing advanced framework SW-ICR-CCR~\cite{cai2019exploring} is not satisfactory. However, BFDA achieves significantly improved results, outperforming the current SOTA methods and approaching the accuracy of Oracle, which highlights the effectiveness of BFDA in weather adaptation scenarios.  Furthermore,  our one-stage-detector-based BFDA is more than six times faster than those based on two-stage detectors, demonstrating our framework's efficiency once again.

\renewcommand\arraystretch{1}
\begin{table*}[]
\begin{center}
\caption{A description of the composition of different frameworks for ablation experiments. The modules have an order in which they are added. This is because the Background Decoupling Module requires the Feature Generation Module to construct the loss function. After the background features are decoupled, we can use the long-short-range domain discriminator to give different attention to backgrounds in different ranges.}
\label{table:abb} 
\begin{tabular}{c|c|c|c|c}
\toprule
          & Background   Decoupling Module & Feature Generation Module & Short-range Discriminator & Long-range Discriminator \\ \hline
BFDA$_{L}$   &                                &                           &                                  & \checkmark                               \\ 
BFDA$_{LF}$  &                                & \checkmark                         &                                  & \checkmark                               \\ 
BFDA$_{LBF}$ & \checkmark                              & \checkmark                         &                                  & \checkmark                               \\ 
BFDA      & \checkmark                              & \checkmark                         & \checkmark                                & \checkmark                               \\ \bottomrule
\end{tabular}
\end{center}
\end{table*}
\renewcommand\arraystretch{1}

\begin{table}
\begin{center}
\caption{scene adaptation: \emph{CityPersons$\rightarrow$Caltech} (FPS on \textbf{V100})}
\label{table:scen}
\renewcommand\arraystretch{1.2}
\setlength{\tabcolsep}{1.9mm}{\scalebox{1}{
\begin{tabular}{c|cccc|c}
\toprule
\multirow{2}{*}{Method} & \multicolumn{4}{c|}{$M R^{-2} (\%)$ ↓} & \multirow{2}{*}{FPS} \\ \cline{2-5}
 & \multicolumn{1}{c|}{\textbf{reasonable}} & \multicolumn{1}{c|}{bare} & \multicolumn{1}{c|}{partial} & heavy &  \\ \hline\hline
Source-only & \multicolumn{1}{c|}{15.91} & \multicolumn{1}{c|}{15.67} & \multicolumn{1}{c|}{18.51} & 31.41 & 217.4 \\ \hline
SCDA~\cite{zhu2019adapting} & \multicolumn{1}{c|}{28.93} & \multicolumn{1}{c|}{28.93} & \multicolumn{1}{c|}{-} & - & 16.7 \\
DAFR~\cite{chen2018domain} & \multicolumn{1}{c|}{18.42} & \multicolumn{1}{c|}{18.42} & \multicolumn{1}{c|}{-} & - & 12.0 \\
SAN~\cite{jiao2021san} & \multicolumn{1}{c|}{14.27} & \multicolumn{1}{c|}{14.27} & \multicolumn{1}{c|}{-} & - & 17.1 \\ \hline
BFDA$_{L}$ & \multicolumn{1}{c|}{9.40} & \multicolumn{1}{c|}{9.30} & \multicolumn{1}{c|}{\textbf{9.29}} & \textbf{24.77} & \multirow{4}{*}{\textbf{217.4}} \\
BFDA$_{LF}$ & \multicolumn{1}{c|}{8.83} & \multicolumn{1}{c|}{8.57} & \multicolumn{1}{c|}{13.40} & 25.38 &  \\
BFDA$_{LBF}$& \multicolumn{1}{c|}{8.29} & \multicolumn{1}{c|}{8.05} & \multicolumn{1}{c|}{12.66} & 25.00 &  \\
BFDA& \multicolumn{1}{c|}{\textbf{7.30}} & \multicolumn{1}{c|}{\textbf{7.02}} & \multicolumn{1}{c|}{11.82} & 25.82 &  \\ \hline
Oracle(Train-on-target) & \multicolumn{1}{c|}{5.38} & \multicolumn{1}{c|}{5.05} & \multicolumn{1}{c|}{0.00} & 38.37 & 217.4 \\ \bottomrule
\end{tabular}}}
\end{center}
\end{table}

\begin{table}
\begin{center}
\caption{weather adaptation: \emph{CityPersons$\rightarrow$Foggy Cityscapes} (FPS on \textbf{V100})}
\label{table:weat} 
\renewcommand\arraystretch{1.2}
\setlength{\tabcolsep}{1.9mm}{\scalebox{1}{
\begin{tabular}{c|cccc|c}
\toprule
\multirow{2}{*}{Method} & \multicolumn{4}{c|}{$M R^{-2} (\%)$ ↓} & \multirow{2}{*}{FPS} \\ \cline{2-5}
 & \multicolumn{1}{c|}{\textbf{reasonable}} & \multicolumn{1}{c|}{bare} & \multicolumn{1}{c|}{partial} & heavy &  \\ \hline\hline
Source-only & \multicolumn{1}{c|}{26.64} & \multicolumn{1}{c|}{19.75} & \multicolumn{1}{c|}{27.13} & 54.35 & 42.7 \\ \hline
DAFR~\cite{chen2018domain} & \multicolumn{1}{c|}{54.71} & \multicolumn{1}{c|}{54.71} & \multicolumn{1}{c|}{-} & - & 5.7 \\
SW-ICR-CCR~\cite{cai2019exploring} & \multicolumn{1}{c|}{49.54} & \multicolumn{1}{c|}{37.95} & \multicolumn{1}{c|}{55.13} & 89.69 & 6.3 \\ \hline
BFDA$_{L}$& \multicolumn{1}{c|}{23.92} & \multicolumn{1}{c|}{16.73} & \multicolumn{1}{c|}{24.85} & 52.53 & \multirow{4}{*}{\textbf{42.7}} \\
BFDA$_{LF}$& \multicolumn{1}{c|}{24.58} & \multicolumn{1}{c|}{18.00} & \multicolumn{1}{c|}{26.37} & 57.28 &  \\
BFDA$_{LBF}$& \multicolumn{1}{c|}{20.62} & \multicolumn{1}{c|}{14.95} & \multicolumn{1}{c|}{20.99} & \textbf{50.95} &  \\
BFDA& \multicolumn{1}{c|}{\textbf{18.57}} & \multicolumn{1}{c|}{\textbf{12.84}} & \multicolumn{1}{c|}{\textbf{19.35}} & 52.21 &  \\ \hline
Oracle(Train-on-target) & \multicolumn{1}{c|}{14.33} & \multicolumn{1}{c|}{9.17} & \multicolumn{1}{c|}{14.32} & 44.22 &  42.7\\ \bottomrule
\end{tabular}}}
\end{center}
\end{table}

\renewcommand\arraystretch{1.3}
\begin{table*}[t]
\begin{center}
\caption{ Generalization evaluation: \emph{Cityscapes$\rightarrow$Foggy Cityscapes}. \textbf{We use framework-level comparison because of the non-portability of the various methods.} ($^\dag$: YOLOv5-based; $^\ddag$: Faster RCNN (vgg16)-based). [\textbf{\underline{Note}} that our BFDA is designed for  one-stage detectors (like YOLOv5). We include experiments with Faster RCNN here just for a fair comparison, but most of the existing Faster RCNN-based SOTA methods cannot be used on  one-stage detectors.]}
\label{table:all_class}
\setlength{\tabcolsep}{4mm}{\scalebox{1}{
\begin{tabular}{c|c|cccccccc}
\toprule
Method & $\operatorname{mAP}(\%)$↑ & person & rider & car & truck & bus & train & mcycle & bicycle \\ \hline\hline
Source-only$^\ddag$ & 25.8 & 33.7 & 35.2 & 13.0 & 28.2 & 9.1 & 18.7 & 31.4 & 24.4 \\
Source-only$^\dag$ & 46.0 & 55.0 & 58.3 & 63.9 & 30.1 & 37.9 & 28.1 & 44.8 & 49.8 \\ \hline\hline
MeGA-CDA~\cite{vs2021mega}$^\ddag$ & 41.8 & 37.7 & 49.0 & 52.4 & 25.4 & 49.2 & 46.9 & 34.5 & 39.0 \\ 
UMT~\cite{deng2021unbiased}$^\ddag$ & 41.7 & 33.0 & 46.7 & 48.6 & 34.1 & 56.5 & 46.8 & 30.4 & 37.3 \\ 
HTCN~\cite{chen2020harmonizing}$^\ddag$ & 39.8 & 33.2 & 47.5 & 47.9 & 31.6 & 47.4 & 40.9 & 32.3 & 37.1 \\
CRDA~\cite{xu2020exploring}$^\ddag$ & 37.4 & 32.9 & 43.8 & 49.2 & 27.2 & 45.1 & 36.4 & 30.3 & 34.6 \\
SWDA~\cite{saito2019strong}$^\ddag$ & 34.3 & 29.9 & 42.3 & 43.5 & 24.5 & 36.2 & 32.6 & 30.0 & 35.3 \\ \hline

CADA~\cite{hsu2020every}$^\dag$ & 40.2 & 41.5 & 43.6 & 57.1 & 29.4 & 44.9 & 39.7 & 29.0 & 36.1\\

SSAL~\cite{munir2021ssal}$^\dag$ & 39.6 & 45.1 & 47.4 & 59.4 & 24.5 & 50.5 & 25.7 & 26.0 & 38.7\\

S-DAYOLO~\cite{li2022cross}$^\dag$ & 39.0 & 42.6 & 42.1 & 61.9 & 23.5 & 40.5 & 39.5 & 24.4 & 37.3\\
DA-YOLO~\cite{zhang2021domain}$^\dag$ & 36.1 & 29.5 & 27.7 & 46.1 & 9.1 & 28.2 & 4.5 & 12.7 & 24.8\\ \hline
\hline

BFDA(Ours)$^\ddag$ & 41.4 & 41.4 & 48.1 & 60.5 & 27.2 & 47.9 & 32.6 & 31.8 & 41.9 \\
BFDA(Ours)$^\dag$ & \textbf{58.1} & \textbf{64.2} & \textbf{65.3} & \textbf{74.2} & \textbf{38.8} & \textbf{62.2} & \textbf{51.8} & \textbf{50.6} & \textbf{58.1} \\\hline\hline
Oracle$^\ddag$ & 43.5 & 37.2 & 48.3 & 52.7 & 35.2 & 52.2 & 48.5 & 35.3 & 38.8  \\ 
Oracle$^\dag$ & 66.4 & 71.4 & 73.6 & 83.3 & 51.6 & 72.8 & 61.4 & 56.9 & 60.2 \\ \bottomrule
\end{tabular}}}
\end{center}
\end{table*}
\renewcommand\arraystretch{1}

\begin{figure*}[t]
\hsize=\textwidth
\centering
\includegraphics[width=1\linewidth]{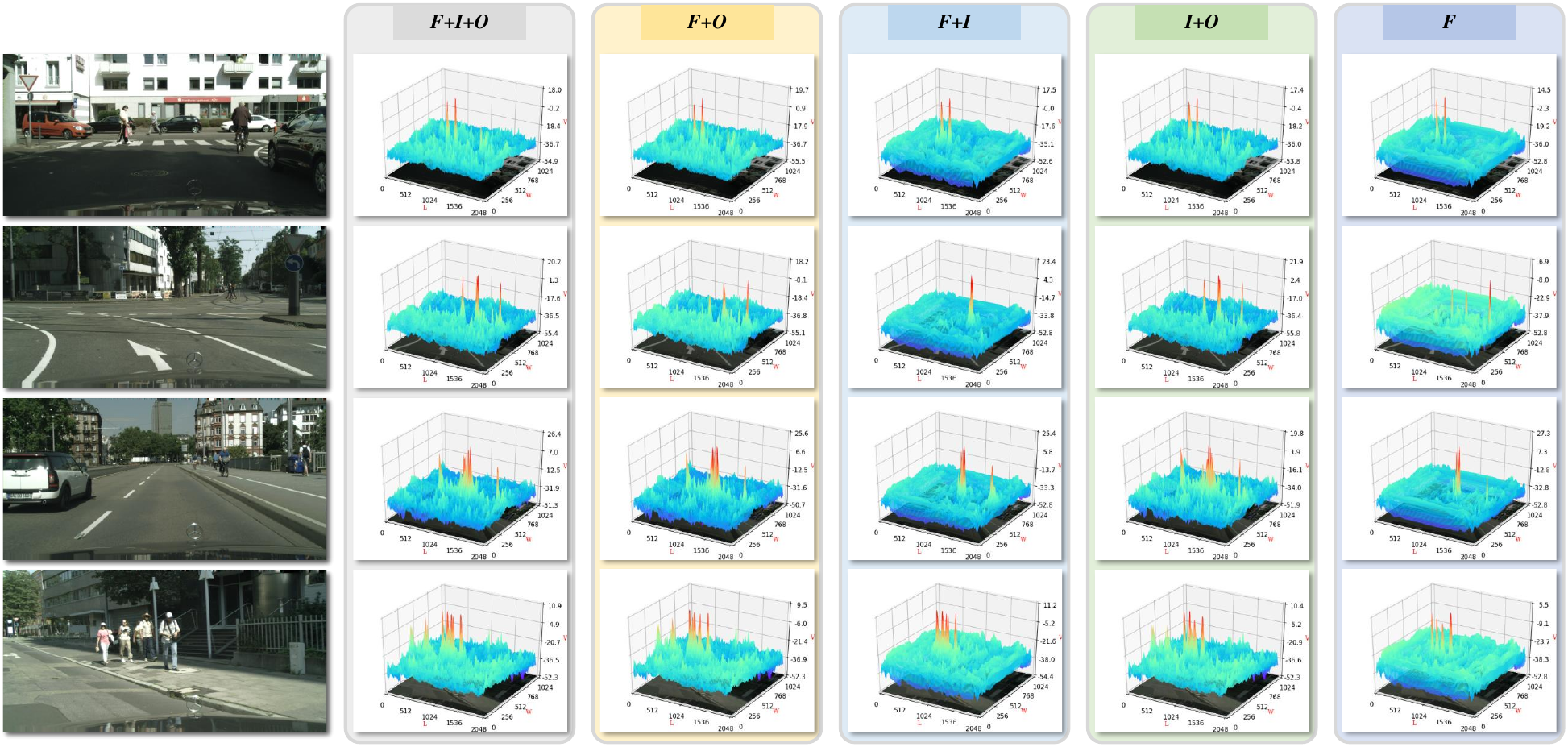}
\caption{Ablation study of the foreground-background feature coupling (YOLOv5-based). $F$, $I$, and $O$ refer to Foreground, Inner-bounding-box background, and Outer-bounding-box background, respectively. Each column in the figure represents the feature maps generated by retaining only a specific part of the original image. For example, the second column ($F+O$) indicates that this column is the corresponding feature map after removing the Inner-bounding-box background ($I$) from the original images. Moreover, we \textbf{cannot} estimate the quality of our background decoupling by visualizing the feature maps generated by BDM (even if all are backgrounds, the feature maps still have some peaks).}
  \label{fig:fig7}
\end{figure*}

\begin{figure}[h]
	\centering
	\subfigure[Original image (Cityscapes)]{
		\begin{minipage}[b]{0.46\linewidth}
		\centering
			\includegraphics[height=0.5\textwidth]{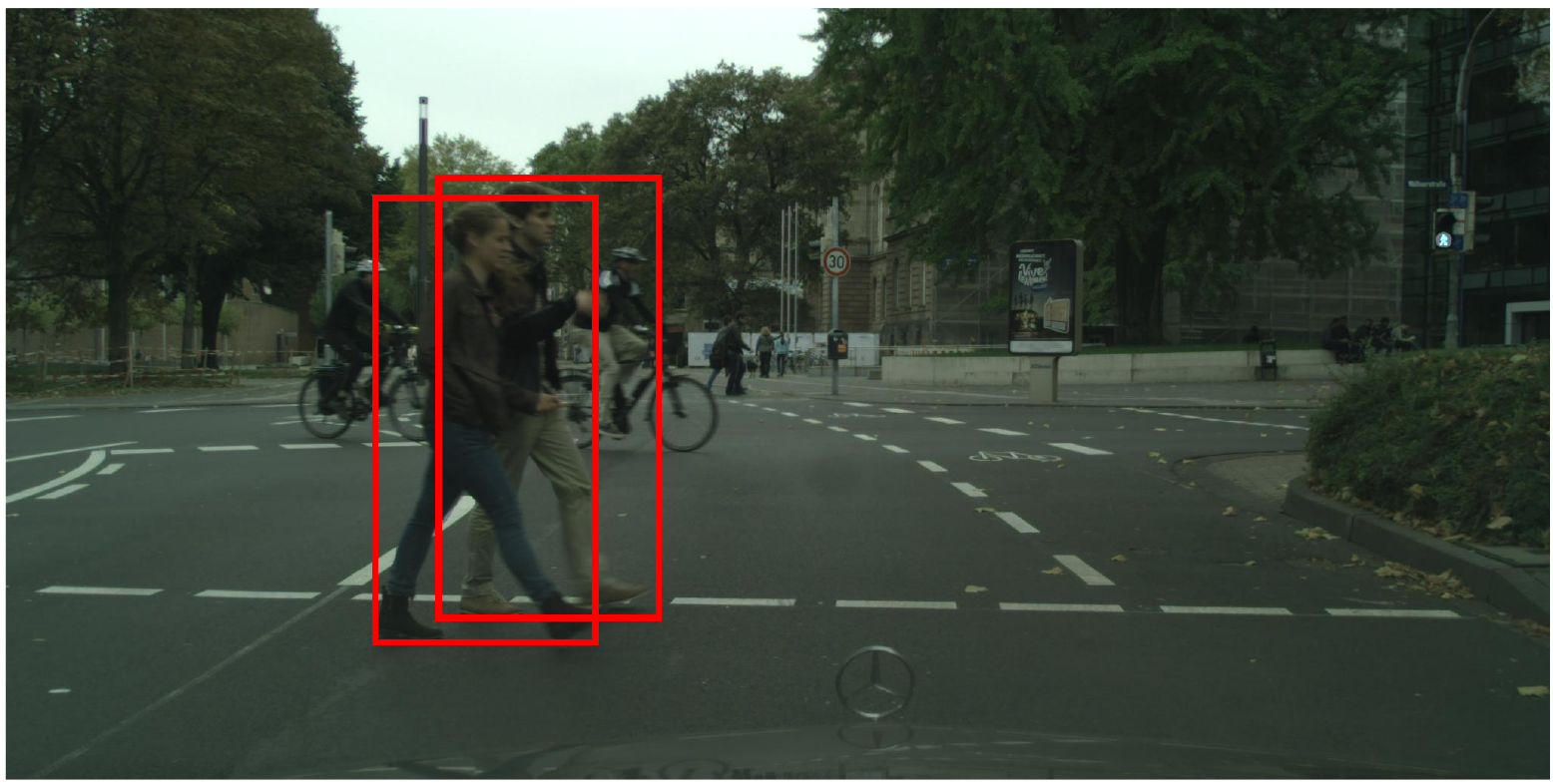} 
		\end{minipage}
		\label{fig:grid_4figs_1cap_4subcap_1}
	}
    	\subfigure[Output of FGM (Cityscapes)]{
    		\begin{minipage}[b]{0.46\linewidth}
    		\centering
   		 	\includegraphics[height=0.5\textwidth]{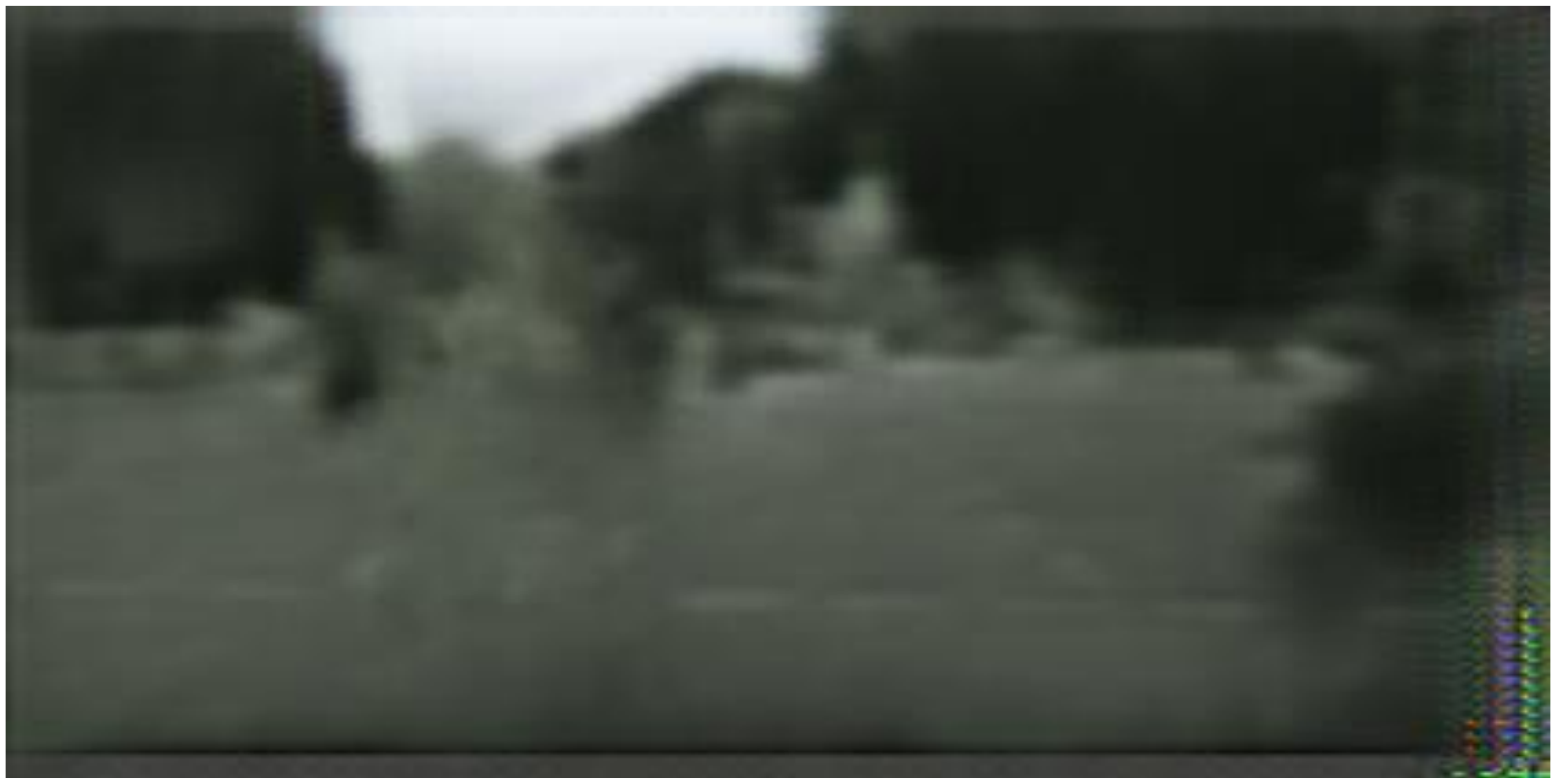}
    		\end{minipage}
		\label{fig:grid_4figs_1cap_4subcap_2}
    	}
    \caption{Background decoupling visualization.\textit{CityPersons$\rightarrow$Caltech}}
	\label{fig:B_v}
\end{figure}

\begin{figure*}[h]
\centering
\includegraphics[width=1\linewidth]{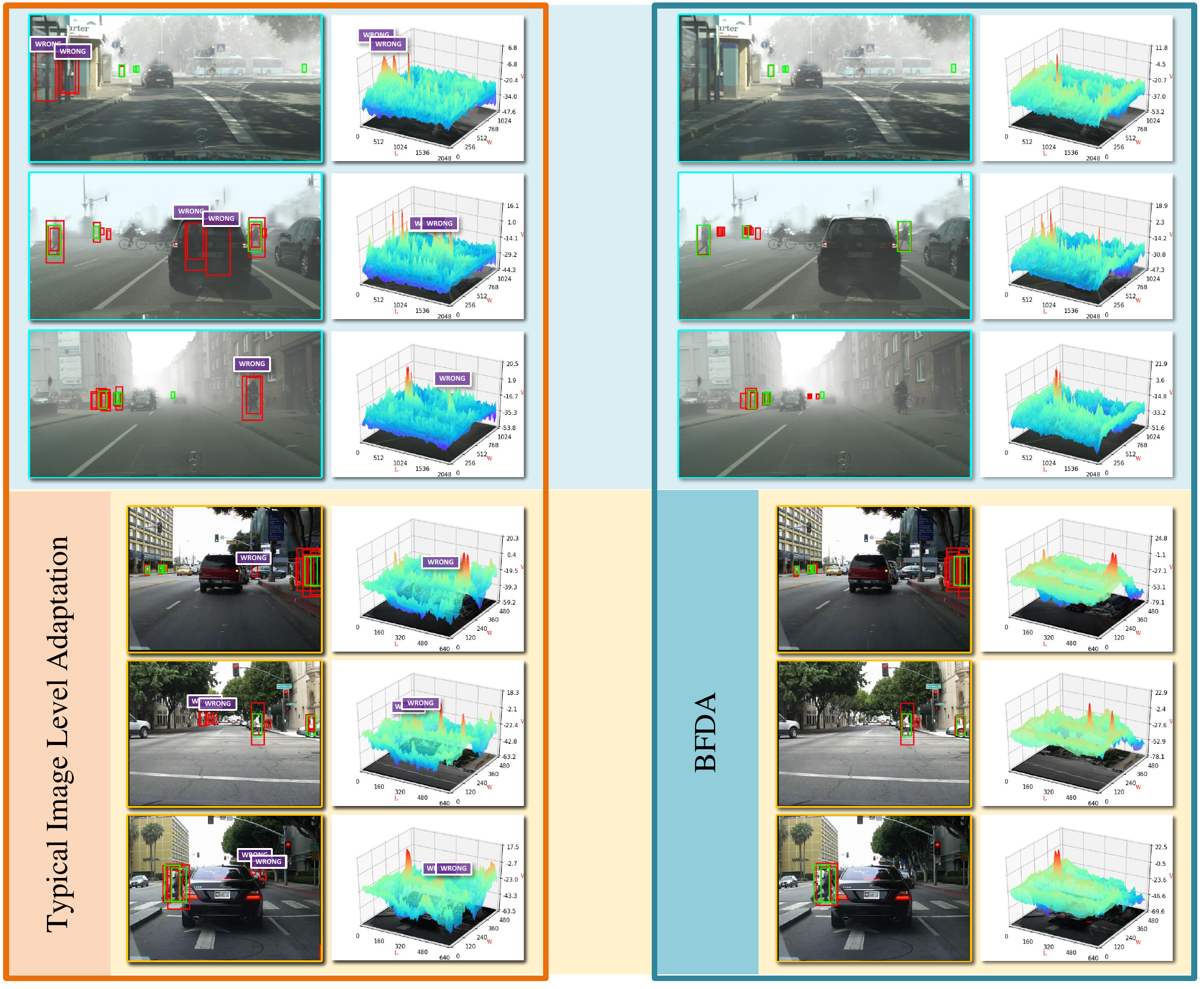}
\caption{Cross-domain pedestrian detection results comparison on \textit{CityPersons$\rightarrow$Foggy Cityscapes} (first three lines)  and \textit{CityPersons$\rightarrow$Caltech}(last three lines). The green boxes are ground truths, and the red boxes are prediction boxes with confidence scores greater than 0.01 before Non-Maximum Suppression~\cite{girshick2014rich}. Purple boxes (marked with "WRONG") indicate wrong prediction bounding boxes.}
  \label{fig8}
\end{figure*}

\subsection{Empirical Analysis}
\label{sec5.4}

$M R^{-2} (reasonable)$ (represented as $M R^{-2} (r)$ for brevity) is the \textbf{most crucial metric} because there are far \textbf{more reasonable pedestrians} than other kinds (bare, partial, heavy) in Caltech and Foggy Cityscapes.

\textbf{Ablation Study: } To comprehensively evaluate the effectiveness of BFDA, ablation studies were conducted by introducing three additional variants of BFDA. The modules were sequentially added in a specific order, with the Feature Generation Module preceding the Background Decoupling Module, as the latter requires the former to construct the loss function. Once the background features are successfully decoupled, the long-short-range domain discriminator can then be employed to assign varying attention to backgrounds within different ranges.

Specifically, based on our base detector YOLOv5, \textbf{BFDA$_L$} only takes the long-range discriminator (transformer) and performs typical image-level domain adaptation. \textbf{BFDA$_{LF}$}  then integrates the Feature Generation Module (+\textit{FGM}) based on BFDA$_L$. \textbf{BFDA$_{LBF}$} further introduces a Background Decoupling Module (+\textit{BDM}) based on BFDA$_{LF}$. Finally, \textbf{BFDA }(our full framework) introduces a long-short-range discriminator (+\textit{LSD}) on top of \textbf{BFDA$_{LBF}$} (as shown in Table \ref{table:abb}).  The results are presented in both  Table \ref{table:scen} and Table \ref{table:weat}, and the following conclusions can be drawn:

(1) \emph{Effects of the Long-range Domain Discriminator}: 
To establish that our approach is not solely reliant on borrowing the Transformer for accuracy improvement, we utilize the long-range discriminator (Transformer only) as our \textbf{baseline} and use the typical image-level adaptation method to achieve domain adaptation. \textbf{BFDA$_L$} generally outperforms source-only results. However, the improvement is limited, suggesting that the commonly used image-level domain adaptation is far from sufficient. There are only $L_{det}$ and $L_{dis}$ here (no $L_{gen}$).

(2) \emph{Effects of the Feature Generation Module}: \textbf{{BFDA$_{LF}$}} employs the Feature Generation Module to force the feature map to contain only background features. However, the experimental results do not seem to be very satisfactory: in the scene adaptation task, the module successfully reduces $M R^{-2} (r)$ from 9.40\% to 8.83\%, while in the weather adaptation task, $M R^{-2} (r)$ increases from 23.92\% to 24.58\%. The main reason is that directly suppressing the foreground features on the original feature map using $L{gen}$ leads to a decline in detection accuracy since the foreground features are essential for detection. The loss function $L_{det}$ promotes the feature map to contain foreground features, while $L_{gen}$ promotes the feature map to contain no foreground features, making the two losses incompatible for simultaneous optimization.

(3) \emph{Effects of the Background Decoupling Module}: \textbf{BFDA$_{LBF}$} adopts a Background Decoupling Module, which takes the original feature map as input and extracts background features with $L_{g e n}$'s help. As shown in Table \ref{table:scen} and Table \ref{table:weat}, \textbf{BFDA${LBF}$} demonstrates further enhancement in cross-domain performance, particularly in the weather adaptation scenario, with a performance gain of over 3\%. This improvement may be attributed to the fact that the addition of the Background Decoupling Module allows $L{gen}$ to primarily train the newly introduced module instead of YOLOv5, potentially mitigating the conflict between $L_{det}$ and $L_{gen}$. 

(4) \emph{Effects of the Long-short-range Domain Discriminator}: The aforementioned modules have successfully decoupled the background feature from the original feature map, and the next step is to use the domain discriminator to analyze the background features. Table \ref{table:dif_range} demonstrates that different ranges of backgrounds have different levels of importance, and short-range backgrounds are more important and should be focused on. Our complete framework \textbf{({BFDA})} has greatly improved over BFDA$_L$: in the scene adaptation task, $M R^{-2} (r)$ reduces from 9.40\% to 7.71\%, and in the weather adaptation task, it reduces from 23.92\% to 18.57\%. We attribute this success to the long-short-range domain discriminator, which combines global and local attention capabilities to effectively analyze complex backgrounds. 

\subsection{Visual Analysis}\label{sec_v}
\textbf{Background decoupling visualization:}  To showcase the decoupling of the background, we present the visualizations of the output images obtained from the \textit{FGM} in Figure \ref{fig:B_v}. Thus, if the generated map exclusively consists of background, the input feature map should only contain background features. As evident from the visualizations, the output of the \textit{FGM} captures almost all the background features. Note that visualizing the feature maps generated by the \textit{BDM} (Figure \ref{fig:fig7}) does not clearly demonstrate the effect of background decoupling.

\textbf{The coupling of foreground and background features in the feature map:} The Background Decoupling Module (\textit{BDM}) and the Feature Generation Module (\textit{FGM}) are proposed to decouple background features from the original feature maps. This section presents a discussion of our third finding: background features and foreground features are completely coupled in the feature maps, implying the necessity of \textit{BDM} and \textit{FGM}.

Ablation experiments were conducted using YOLOv5 on the Citypersons dataset to investigate the impact of three different regions (outer-bounding-box background, inner-bounding-box background, and foreground) on the original feature maps of YOLOv5. As shown in Figure~\ref{fig:fig7}, the visualization results reveal an interesting phenomenon: \textbf{the peaks} on each feature map corresponding to the foreground region of the original image are primarily generated by background information. One possible reason for this is that the detection network often models the spatial contextual relations between the foreground and background regions. Upon removal of the background information, these "foreground" peaks are significantly weakened; however, removing the foreground information does not result in significant changes to these "foreground" peaks.

\textbf{Visualization of the detection results:}
We visualize the detection results (Figure~\ref{fig8}) on the scene adaptation task (\textit{CityPersons$\rightarrow$Caltech}) and the weather adaptation task (\textit{CityPersons$\rightarrow$Foggy Cityscapes}). The visualization results clearly demonstrate that the background-focused distribution alignment has much fewer wrong predictions than the typical image-level adaptation, and further verifies the foreground-background feature mismatch issue (the main reason for the wrong prediction boxes) proposed in the main text.

\subsection{Generalization Evaluation.}
We have conducted comprehensive analyses of BFDA on the cross-domain pedestrian detection task and obtained promising results. Our key finding is that background inconsistency dominates the domain gap, an insight that may also be applicable to other detection tasks and that may have broader implications. Based on this insight, we extended BFDA to perform experiments on the general object detection task, simultaneously treating all classes as foreground. The results are presented in Table \ref{table:all_class}, where we compare BFDA with several SOTA domain adaptation methods based on Faster RCNN. As shown, BFDA outperforms almost all other SOTA frameworks \cite{wang2021domain, rezaeianaran2021seeking, vs2021mega, deng2021unbiased, chen2020harmonizing, xu2020exploring, saito2019strong}, thereby validating the effectiveness of BFDA for adapting general object detectors.

\textbf{Note} that our BFDA is designed for  one-stage detectors (such as YOLOv5) because  one-stage detectors can only use image-level cross-domain and must face this issue. While instance-level cross-domain adaptation can partially address this issue by allowing the Region Proposal Network (RPN) to propose foregrounds individually for alignment, we include experiments with Faster RCNN for a fair comparison, as most existing state-of-the-art (SOTA) methods are based on Faster RCNN.  It is not feasible to directly apply existing two-stage detector-based SOTA methods to  one-stage detectors, as instance-free one-stage detectors lack the necessary conditions for instance-level cross-domain adaptation. 

\color{black}

\subsection{Loss Function Coefficient Analysis}
As shown in Equation~\ref{eq:tl}, the loss function of BFDA is $L=\alpha \cdot L_{d e t}+\beta \cdot L_{g e n}+\gamma \cdot L_{d i s}$. The coefficients $\alpha$, $\beta$, and $\gamma$ have a great influence on the results. To demonstrate this effect, we present the experimental results of coefficient analysis under the \emph{CityPersons $\rightarrow$Caltech} setting in Table~\ref{table:ls_ca}. 

From Table~\ref{table:ls_ca}, we can make the following conclusions:
\begin{itemize}
    \item Exp. 1-4 show that the weight $\gamma$ of the discriminator loss item should be set smaller than the other two coefficients since a larger $\gamma \cdot L_{d i s}$ may worsen the training instability due to the adversarial process, causing “mode collapse” that makes the training not converge.
    \item In Exp. 5-12, with $\gamma$ varying from 5e-2 to 1e-4, superior performance can be achieved when $\gamma$ is approximately 1e-2. This is because too large $\gamma$ will cause "mode collapse", and too small $\gamma$ will lead to invalid cross-domain adversarial training. Therefore, a mediate gamma is set to achieve optimal performance.
    \item In Exp. 7 and 13-16, with the weight of the generator loss item $\beta$ varying from 1 to 1e-3 and $\gamma$ set as 1e-2, superior performance can be achieved when $\beta$ is approximately 1e-1. Under this setting, we can approach a better trade-off between the cross-domain adversarial training and the extraction of the background information from the detector training.
\end{itemize}

\renewcommand\arraystretch{1.1}
\begin{table}[t]
\begin{center}
\caption{Loss function coefficient analysis: \emph{Citypersons$\rightarrow$Caltech}.}
\label{table:ls_ca}
\setlength{\tabcolsep}{1.8mm}{\scalebox{1}{
\begin{tabular}{c|ccc|cccc}
\toprule
\multirow{2}{*}{\begin{tabular}[c]{@{}c@{}}Index\end{tabular}} & \multirow{2}{*}{$\alpha$} & \multirow{2}{*}{$\beta$} & \multirow{2}{*}{$\gamma$} & \multicolumn{4}{c}{$M R^{-2} (\%)$ ↓}                                                                           \\ \cline{5-8} 
                                                                             &                           &                          &                           & \multicolumn{1}{c|}{\textbf{reasonable}} & \multicolumn{1}{c|}{bare}  & \multicolumn{1}{c|}{partial} & heavy \\ \hline\hline
1                                                                            & 1                         & 1                        & 1                         & \multicolumn{1}{c|}{11.17}      & \multicolumn{1}{c|}{10.97} & \multicolumn{1}{c|}{13.82}   & 27.48 \\ \hline
2                                                                            & 1                         & 1                        & 0.1                       & \multicolumn{1}{c|}{8.51}       & \multicolumn{1}{c|}{7.94}  & \multicolumn{1}{c|}{19.26}   & 30.11 \\ \hline
3                                                                            & 0.1                       & 1                        & 1                         & \multicolumn{1}{c|}{11.16}      & \multicolumn{1}{c|}{10.91} & \multicolumn{1}{c|}{14.14}   & 27.77 \\ \hline
4                                                                            & 1                         & 0.1                      & 1                         & \multicolumn{1}{c|}{11.09}      & \multicolumn{1}{c|}{10.87} & \multicolumn{1}{c|}{13.37}   & 27.35 \\ \hline\hline
5                                                                           & 1                         & 1                        & 0.05                      & \multicolumn{1}{c|}{8.07}       & \multicolumn{1}{c|}{7.68}  & \multicolumn{1}{c|}{16.39}   & 31.63 \\ \hline
6                                                                            & 1                         & 1                        & 0.02                      & \multicolumn{1}{c|}{8.07}       & \multicolumn{1}{c|}{8.03}  & \multicolumn{1}{c|}{14.33}   & 28.80  \\ \hline
7                                                                            & 1                         & 1                        & 0.01                      & \multicolumn{1}{c|}{7.59}       & \multicolumn{1}{c|}{7.38}  & \multicolumn{1}{c|}{\textbf{10.78}}   & 26.19 \\ \hline
8                                                                            & 1                         & 1                        & 0.005                     & \multicolumn{1}{c|}{8.79}       & \multicolumn{1}{c|}{8.25}  & \multicolumn{1}{c|}{12.11}   & 27.53 \\ \hline
9                                                                           & 1                         & 1                        & 0.002                     & \multicolumn{1}{c|}{8.91}       & \multicolumn{1}{c|}{8.66}  & \multicolumn{1}{c|}{12.21}   & 25.98 \\ \hline
10                                                                            & 1                         & 1                        & 0.001                     & \multicolumn{1}{c|}{8.55}       & \multicolumn{1}{c|}{8.29}  & \multicolumn{1}{c|}{11.78}   & 26.07 \\ \hline
11                                                                           & 1                         & 1                        & 0.0005                    & \multicolumn{1}{c|}{8.80}       & \multicolumn{1}{c|}{8.64}  & \multicolumn{1}{c|}{11.72}   & \textbf{25.63} \\ \hline
12                                                                           & 1                         & 1                        & 0.0001                    & \multicolumn{1}{c|}{9.00}       & \multicolumn{1}{c|}{8.81}  & \multicolumn{1}{c|}{12.38}   & 26.77 \\ \hline\hline
13                                                                           & 1                         & 0.1                      & 0.01                      & \multicolumn{1}{c|}{\textbf{7.30}}       & \multicolumn{1}{c|}{\textbf{7.02}}  & \multicolumn{1}{c|}{11.82}   & 25.82 \\ \hline
14                                                                           & 1                         & 0.05                     & 0.01                      & \multicolumn{1}{c|}{7.55}       & \multicolumn{1}{c|}{7.12}  & \multicolumn{1}{c|}{14.56}   & 27.55 \\ \hline
15                                                                           & 1                         & 0.01                     & 0.01                      & \multicolumn{1}{c|}{8.66}       & \multicolumn{1}{c|}{8.32}  & \multicolumn{1}{c|}{15.07}   & 28.12 \\ \hline
16                                                                           & 1                         & 0.001                    & 0.01                      & \multicolumn{1}{c|}{9.30}       & \multicolumn{1}{c|}{9.09}  & \multicolumn{1}{c|}{12.55}   & 26.93 \\ \bottomrule
\end{tabular}

}}
\end{center}
\end{table}
\renewcommand\arraystretch{1}

\subsection{More Ablation Experiments}
To better demonstrate the respective roles of each module we designed, we show all possible ablation experiments based on the \emph{CityPersons $\rightarrow$Caltech} setting, as shown in Table~\ref{table:AE}. \textit{BDM}, \textit{FGM}, \textit{SD}, \textit{LD} in the table represent the Background Decoupling Module, Feature Generation Module, Short-range Discriminator and Long-range Discriminator, respectively.

\renewcommand\arraystretch{1.2}
\begin{table}[t]
\begin{center}
\caption{More Ablation Experiments: \emph{Citypersons$\rightarrow$Caltech}.}
\label{table:AE}
\setlength{\tabcolsep}{1.2mm}{\scalebox{1}{

\begin{tabular}{c|cccc|cccc}
\toprule
\multirow{2}{*}{Method} & \multirow{2}{*}{\textit{BDM}} & \multirow{2}{*}{\textit{FGM}} & \multirow{2}{*}{\textit{SD}} & \multirow{2}{*}{\textit{LD}} & \multicolumn{4}{c}{$M R^{-2} (\%)$ ↓}                                                                          \\ \cline{6-9} 
                        &                      &                      &                     &                     & \multicolumn{1}{c|}{\textbf{reasonable}} & \multicolumn{1}{c|}{bare} & \multicolumn{1}{c|}{partial} & heavy \\ \hline\hline
BFDA$_L$                 &                      &                      &                     & \checkmark                   & \multicolumn{1}{c|}{9.40}        & \multicolumn{1}{c|}{9.30}  & \multicolumn{1}{c|}{\textbf{9.29}}    & \textbf{24.77} \\ \hline
BFDA$_{LS}$                &                      &                      & \checkmark                   & \checkmark                   & \multicolumn{1}{c|}{8.46}       & \multicolumn{1}{c|}{8.32} & \multicolumn{1}{c|}{11.03}   & 26.19 \\ \hline
BFDA$_{LF}$               &                      & \checkmark                    &                     & \checkmark                   & \multicolumn{1}{c|}{8.83}       & \multicolumn{1}{c|}{8.57} & \multicolumn{1}{c|}{13.40}    & 25.38 \\ \hline
BFDA$_{LB}$                & \checkmark                    &                      &                     & \checkmark                   & \multicolumn{1}{c|}{9.12}       & \multicolumn{1}{c|}{8.86} & \multicolumn{1}{c|}{13.90}    & 26.78 \\ \hline
BFDA$_{LBF}$               & \checkmark                    & \checkmark                    &                     & \checkmark                   & \multicolumn{1}{c|}{8.29}       & \multicolumn{1}{c|}{8.05} & \multicolumn{1}{c|}{12.66}   & 25.00    \\ \hline
BFDA$_{LSF}$               &                      & \checkmark                    & \checkmark                   & \checkmark                   & \multicolumn{1}{c|}{8.37}       & \multicolumn{1}{c|}{8.08} & \multicolumn{1}{c|}{12.50}    & 27.62 \\ \hline
BFDA$_{LSB}$               & \checkmark                    &                      & \checkmark                   & \checkmark                   & \multicolumn{1}{c|}{8.78}       & \multicolumn{1}{c|}{8.52} & \multicolumn{1}{c|}{14.08}   & 26.02 \\ \hline
BFDA                    & \checkmark                    & \checkmark                    & \checkmark                   & \checkmark                   & \multicolumn{1}{c|}{\textbf{7.30}}        & \multicolumn{1}{c|}{\textbf{7.02}} & \multicolumn{1}{c|}{11.82}   & 25.82 \\ \bottomrule
\end{tabular}

}}
\end{center}
\end{table}
\renewcommand\arraystretch{1}

We can make the following analysis:
\begin{itemize}
    \item \textbf{BFDA$_L$} serves as the baseline with a transformer-based discriminator (long-range).
    \item \textbf{BFDA$_{LS}$} improves the baseline by adding the HRNet branch to the discriminator, forming a complete \textit{LSD}.
    \item \textbf{BFDA$_{LF}$} incorporates the \textit{FGM}. As mentioned in Sec.~\ref{sec5.4}, while it can decouple the background, the absence of \textit{BDM} leads to conflicts between $L_{det}$ and $L_{gen}$.
    \item \textbf{BFDA$_{LB}$} incorporates the \textit{BDM}. The lack of a suitable \textit{F}GM prevents $L_{gen}$ from effectively training the \textit{BDM}'s ability to decouple the background.
    \item \textbf{BFDA$_{LBF}$} incorporates \textit{BDM} and \textit{FGM}. This model is the best except for the full model because both \textit{BDM} and \textit{FGM} are reasonably trained. The only shortcoming is that the influence of short-range features is not considered.
    \item \textbf{BFDA$_{LSF}$} incorporates \textit{SD} and \textit{FGM}. Similar to \noindent\textbf{BFDA$_{LF}$}, conflicts between $L_{det}$ and $L_{gen}$ lead to unsatisfactory results.
    \item \textbf{BFDA$_{LSB}$} incorporates \textit{BDM} and \textit{SD}. Similar to  BFDA$_{LB}$, the absence of \textit{FGM} makes \textit{BDM} unable to be trained reasonably, resulting in less-than-perfect results.
    \item \textbf{BFDA} incorporates all components, and this whole model achieves optimal performance.
\end{itemize}

\subsection{Verifying the effectiveness of BFDA on the SSD detector}
Our BFDA is specifically designed for instance-free one-stage detectors and addresses the foreground-background feature misalignment issue. To further verify the effectiveness of BFDA, we conducted experiments using another instance-free one-stage detector, SSD~\cite{liu2016ssd}. First, we verify that background information is also very important for SSD detector, as shown in Table~\ref{table:SSD_1}. In this table, $all$ refers to using normal images, $no\_OB$ refers to removing outer bounding-box background images, $no\_IB$ refers to removing inner bounding-box background images, and $no\_F$ refers to removing foreground images. Additionally, \textit{no\_1.0\_2.0}, \textit{no\_1.5\_2.5}, and \textit{no\_2.0\_3.0} have the same meaning as described in Table~\ref{table:dif_range}. The results align well with our research motivation, indicating that using BFDA is effective.

\renewcommand\arraystretch{1.1}
\begin{table}[t]
\begin{center}
\caption{Verify the sensitivity of SSD detector to changes in background features. The dataset is based on \textit{Cityscapes}.}
\label{table:SSD_1}
\setlength{\tabcolsep}{3.5mm}{\scalebox{1}{

\begin{tabular}{c|cccc}
\toprule
\multirow{2}{*}{Test Type} & \multicolumn{4}{c}{$M R^{-2} (\%)$ ↓}                                                                           \\ \cline{2-5} 
                           & \multicolumn{1}{c|}{\textbf{reasonable}} & \multicolumn{1}{c|}{bare}  & \multicolumn{1}{c|}{partial} & heavy \\ \hline\hline
$all$                        & \multicolumn{1}{c|}{7.92}       & \multicolumn{1}{c|}{4.94}  & \multicolumn{1}{c|}{7.57}    & 15.21 \\ \hline\hline
$no\_OB$                     & \multicolumn{1}{c|}{55.70}       & \multicolumn{1}{c|}{49.88} & \multicolumn{1}{c|}{55.99}   & 72.05 \\ \hline
$no\_IB$                     & \multicolumn{1}{c|}{44.69}      & \multicolumn{1}{c|}{33.57} & \multicolumn{1}{c|}{43.10}    & 72.22 \\ \hline
$no\_F$                      & \multicolumn{1}{c|}{40.17}      & \multicolumn{1}{c|}{30.59} & \multicolumn{1}{c|}{41.19}   & 55.01 \\ \hline\hline
\textit{no\_1.0\_2.0}                 & \multicolumn{1}{c|}{61.21}      & \multicolumn{1}{c|}{57.35} & \multicolumn{1}{c|}{61.11}   & 76.35 \\ \hline
\textit{no\_1.5\_2.5}                & \multicolumn{1}{c|}{36.90}       & \multicolumn{1}{c|}{29.13} & \multicolumn{1}{c|}{38.75}   & 55.55 \\ \hline
\textit{no\_2.0\_3.0}                & \multicolumn{1}{c|}{26.07}      & \multicolumn{1}{c|}{20.70}  & \multicolumn{1}{c|}{25.67}   & 44.08 \\ \bottomrule
\end{tabular}

}}
\end{center}
\end{table}
\renewcommand\arraystretch{1}


 Second, we further designed a framework by applying the BFDA method to SSD. Specifically, we use BDM and FGM to extract background features from the feature map output by the SSD backbone layer and then feed the background features into the LSD domain discriminator. The results shown in Table~\ref{table:SSD_2} indicate that our BFDA method is also effective on SSD, which demonstrates the generalization ability of our method on different image-level detectors.

\renewcommand\arraystretch{1.1}
\begin{table}[t]
\begin{center}
\caption{Verify our BFDA method on SSD: \emph{Citypersons$\rightarrow$Caltech}.}
\label{table:SSD_2}
\setlength{\tabcolsep}{3mm}{\scalebox{1}{

\begin{tabular}{c|cccc}
\toprule
\multirow{2}{*}{Method} & \multicolumn{4}{c}{$M R^{-2} (\%)$ ↓}                                                                           \\ \cline{2-5} 
                        & \multicolumn{1}{c|}{\textbf{reasonable}} & \multicolumn{1}{c|}{bare}  & \multicolumn{1}{c|}{partial} & heavy \\ \hline\hline
Source-only             & \multicolumn{1}{c|}{24.25}      & \multicolumn{1}{c|}{23.43} & \multicolumn{1}{c|}{40.46}   & 50.22 \\ \hline
SSD + BFDA$_L$                 & \multicolumn{1}{c|}{17.57}      & \multicolumn{1}{c|}{17.10} & \multicolumn{1}{c|}{24.62}   & 44.86 \\ \hline
SSD + BFDA                    & \multicolumn{1}{c|}{16.81}      & \multicolumn{1}{c|}{16.02} & \multicolumn{1}{c|}{32.40}   & 43.56 \\ \hline
Oracle                  & \multicolumn{1}{c|}{8.09}       & \multicolumn{1}{c|}{7.36}  & \multicolumn{1}{c|}{20.65}   & 30.79 \\ \bottomrule
\end{tabular}

}}
\end{center}
\end{table}
\renewcommand\arraystretch{1}

\color{black}

\color{black}

\subsection{Experiments on pseudo-label processing methods}
We utilize high-confidence detection boxes obtained from the previous epoch of YOLOv5 to generate pseudo-labels, which are then employed to aid the training of the \textit{BDM} and \textit{FGM}. Specifically, the foreground pseudo-label for an image is generated by taking the average pixel value of the image. Our primary motivation is to reduce the prominence of foreground features by diluting them with the image's average pixel value, as shown in Figure~\ref{fig:B_v}. Using other colors, such as pure black or white, results in an accentuation of foreground features. To substantiate our claim, we present results in Table~\ref{table:PL_1}, where "random" refers to a random value from 0 to 255. 

According to the results in Table~\ref{table:PL_1}, we can make the following analysis:

\begin{itemize}
    \item Utilizing pure black or white as a pseudo-label leads to "abnormal prominence" in the feature map, thereby adversely affecting the outcome of cross-domain adaptation.
    \item The use of random colors as pseudo-labels is not advisable since it also tends to introduce some "abnormal prominence" in the pseudo-label sets.
    \item Choosing the average color value of the images as pseudo-label color is the best option for avoiding any "abnormal prominence" resulting from significant differences between the foreground and background features.
\end{itemize}
In summary, using the average color value as a pseudo-label is the most practical and appropriate choice.

We also try to load the pretrained Deeplabv3~\cite{chen2017rethinking} semantic segmentation model to assist in the generation of pseudo-labels. However, since the semantic segmentation model cannot be trained (otherwise, it means obtaining additional labels, which will lead to unfair comparison), the effect of segmentation is limited, and the results obtained are also limited.

\renewcommand\arraystretch{1.2}
\begin{table}[t]
\begin{center}
\caption{Pseudo-label Analysis: \emph{Citypersons$\rightarrow$Caltech}.}
\label{table:PL_1}
\setlength{\tabcolsep}{3mm}{\scalebox{1}{

\begin{tabular}{c|cccc}
\toprule
\multirow{2}{*}{Pseudo Label} & \multicolumn{4}{c}{$M R^{-2} (\%)$ ↓}                                                                       \\ \cline{2-5} 
                                    & \multicolumn{1}{c|}{reasonable} & \multicolumn{1}{c|}{bare} & \multicolumn{1}{c|}{partial} & heavy \\ \hline\hline
average                             & \multicolumn{1}{c|}{7.30}       & \multicolumn{1}{c|}{7.02} & \multicolumn{1}{c|}{11.82}   & 25.82 \\ \hline\hline
black                                   & \multicolumn{1}{c|}{8.37}       & \multicolumn{1}{c|}{8.11} & \multicolumn{1}{c|}{12.66}   & 28.27 \\ \hline
white                                   & \multicolumn{1}{c|}{8.21}       & \multicolumn{1}{c|}{7.99} & \multicolumn{1}{c|}{12.79}   & 26.15 \\ \hline
random                              & \multicolumn{1}{c|}{8.28}       & \multicolumn{1}{c|}{7.92} & \multicolumn{1}{c|}{13.44}   & 24.80\\
\hline\hline
Deeplabv3 + average                              & \multicolumn{1}{c|}{8.08}       & \multicolumn{1}{c|}{7.77} & \multicolumn{1}{c|}{12.19}   & 27.25\\ \bottomrule
\end{tabular}

}}
\end{center}
\end{table}
\renewcommand\arraystretch{1}

\subsection{Improvements to focus on short-range backgrounds}
We are currently implementing long-short-range attention by leveraging a combination of convolutional neural networks and transformers. Our research is centered around the belief that the convolutional neural network is a model that is inherently rich in local attention.

\begin{figure}[h]
	\centering
	\subfigure[Original image (Cityscapes)]{
		\begin{minipage}[b]{0.46\linewidth}
		\centering
			\includegraphics[height=0.5\textwidth]{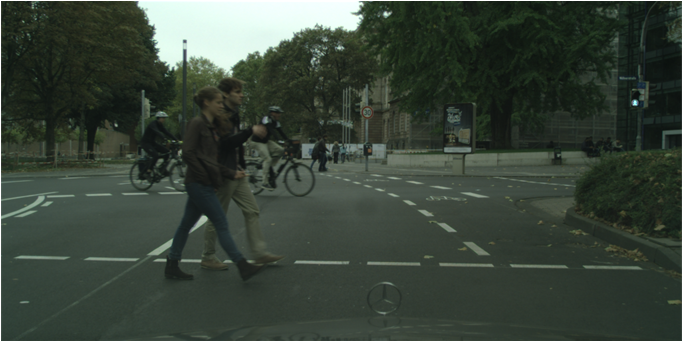} 
		\end{minipage}
		\label{fig:short_improve_1}
	}
    	\subfigure[Discriminator Weight Map]{
    		\begin{minipage}[b]{0.46\linewidth}
    		\centering
   		 	\includegraphics[height=0.5\textwidth]{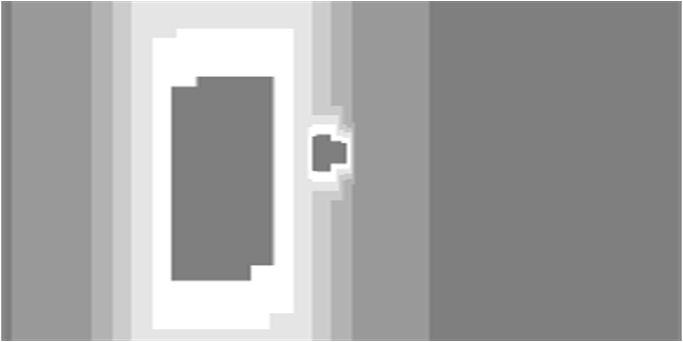}
    		\end{minipage}
		\label{fig:short_improve_2}
    	}
    \caption{Backgrounds of different ranges have different weights.}
	\label{fig:short_improve}
\end{figure}

We further explore a simple yet effective method of assigning varying weights to the background of different ranges, as illustrated in Figure~\ref{fig:short_improve}. Specifically, we assign a weight of 2.0 at 1$\sim$1.5 times the size of the bounding boxes in the background around the instance. 1.5$\sim$2.0 times the background weight is 1.8. 2.0$\sim$2.5 times the background weight is 1.6. 2.5$\sim$3.0 times the background weight is 1.4. 3.0$\sim$5.0 times the background weight is 1.2. The foreground part and the 5.0 times outer background part are assigned a weight of 1.0. Our aim is to optimize the attention mechanism and improve the model's overall performance. This method is indeed effective, and the results are shown in Table~\ref{table:SI}.

However, it should be noted that such a method requires far more computing resources and training time than the original method (specifically, nearly 3 times what is required by the original method). Despite its slightly better results, we have chosen not to use it in this work, as it deviates from our original intention of pursuing a cross-domain pedestrian detector with both high efficiency and good performance.

\renewcommand\arraystretch{1.1}
\begin{table}[t]
\begin{center}
\caption{Improvements on short-range: \emph{Citypersons$\rightarrow$Caltech}.}
\label{table:SI}
\setlength{\tabcolsep}{3mm}{\scalebox{1}{

\begin{tabular}{c|cccc}
\toprule
\multirow{2}{*}{Method}                                               & \multicolumn{4}{c}{$M R^{-2} (\%)$ ↓}                                                                          \\ \cline{2-5} 
                                                                      & \multicolumn{1}{c|}{reasonable} & \multicolumn{1}{c|}{bare} & \multicolumn{1}{c|}{partial} & heavy \\ \hline\hline
BFDA                                                                  & \multicolumn{1}{c|}{7.30}       & \multicolumn{1}{c|}{7.02} & \multicolumn{1}{c|}{11.82}   & 25.82 \\ \hline
\begin{tabular}[c]{@{}c@{}}BFDA + \\ short range improve\end{tabular} & \multicolumn{1}{c|}{7.05}       & \multicolumn{1}{c|}{6.76} & \multicolumn{1}{c|}{12.65}   & 28.51 \\ \bottomrule
\end{tabular}

}}
\end{center}
\end{table}
\renewcommand\arraystretch{1}

\color{black}

\section{Conclusions}
We uncover a problem with the direct application of image-level domain adaptation on instance-free one-stage detectors and investigate cross-domain PD tasks from a new perspective. We also find that mainstream detectors are generally sensitive to background variations, further inspiring us to develop a new background-focused distribution alignment framework BFDA. BFDA comprises three essential parts: the Background Decoupling Module, the Feature Generation Module, and the long-short-range domain discriminator. We conduct extensive experiments on multiple benchmark datasets, and their results clearly show that our BFDA surpasses the existing SOTA frameworks with great advantages in detection accuracy. Meanwhile, as our framework is based on advanced YOLOv5, the inference speed can reach 7$\sim$12 times the FPS of the existing SOTA frameworks.

\section*{Acknowledgement}
This work is supported by National Natural Science Foundation of China (No. 62071127, U1909207),  Shanghai Natural Science Foundation(No. 23ZR1402900), Zhejiang Lab Project (No. 2021KH0AB05).

\vfill
\bibliographystyle{IEEEtran}
\bibliography{TIP_submission_Final_Version}

\end{document}